\documentclass[10pt]{article} 

\usepackage[usenames]{color}
\usepackage[usenames,dvipsnames]{xcolor}
\usepackage{ifthen}

\usepackage{amsfonts}
\usepackage{amsmath}
\usepackage{amsthm}
\usepackage{algorithmic}
\usepackage[boxed]{algorithm}
\usepackage{amssymb}

\usepackage{fullpage}
\usepackage{natbib}

\usepackage{hyperref}
\hypersetup{
  colorlinks   = true, 
  urlcolor     = blue, 
  linkcolor    = blue, 
  citecolor   = blue 
}

\usepackage[hyphenbreaks]{breakurl}

\newtheorem{myfact}{Fact}
\newtheorem{mythm}{Theorem}
\newtheorem{mydef}{Definition}
\newtheorem{mylem}{Lemma}

\newcommand{\chgins}[1]{#1}

\newcommand{\chgred}[1]{{#1}}

\newcommand{\compilehidecomments}{false}

\ifthenelse{ \equal{\compilehidecomments}{true} }{%
	\newcommand{\wei}[1]{}
	\newcommand{\yang}[1]{}
	\newcommand{\yajun}[1]{}
}{
	\newcommand{\wei}[1]{{\color{blue!50!black}  [\text{Wei:} #1]}}
	\newcommand{\yang}[1]{{\color{brown!60!black} [\text{Yang:} #1]}}
	\newcommand{\yajun}[1]{{\color{green!50!black} [\text{Yajun:} #1]}}
}

\newcommand{\sectionspace}{0pt}
\newcommand{\subsectionspace}{0pt}

\newcommand{\calS}{\mathcal{S}}

\newcommand{\calN}{\mathcal{N}}

\newcommand{\E}{\mathbb{E}}
\newcommand{\R}{\mathbb{R}}

\newcommand{\bp}{{\boldsymbol \mu}}
\newcommand{\opt}{{\mathrm{opt}}}
\newcommand{\argmax}{\operatornamewithlimits{argmax}}
\newcommand{\argmin}{\operatornamewithlimits{argmin}}
\newcommand{\bad}{{\mathrm{B}}}
\newcommand{\base}{{\ell}}

\newcommand{\I}{\mathbb{I}}

\newcommand{\trig}[1]{\tilde{#1}}

\begin{document}

\onecolumn

%
%
\title{Combinatorial Multi-Armed Bandit and \\
	Its Extension to Probabilistically Triggered Arms\thanks{A preliminary
	version of this paper appears in ICML'2013 \citep{CWY13}. The current version contains the extension
	of CMAB to accommodate probabilistically triggered arms and its application to
	social influence maximization. The research is partially supported by the National Natural Science
	Foundation of China (Grant No. 61433014).
	}}

%

\author{
Wei Chen\thanks{Contact author: Wei Chen, Microsoft Research Asia, Microsoft Asia R\&D Headquarters,
	Building 2 14-171, 5 Dan Ling Street, Haidian District, Beijing, China, 100080.}\\
Microsoft\\
Beijing, China\\
weic@microsoft.com
\and
Yajun Wang\\
Microsoft\\
Sunnyvale, CA, U.S.A.\\
yajunw@microsoft.com
\and
Yang Yuan\\
Cornell University\\
Ithaca, NY, U.S.A.\\
yangyuan@cs.cornell.edu
\and
Qinshi Wang \\
Tsinghua University \\
Beijing, China \\
wangqinshi1995@gmail.com
}

\date{}

\maketitle

\begin{abstract}
We define a general framework for a large class of combinatorial
	multi-armed bandit (CMAB) problems, where subsets of 
	base arms with unknown distributions
	form {\em super arms}.
In each round, a super arm is played and the base arms contained in the super arm
	are played and their outcomes are observed.
We further consider the extension in which more base arms could be probabilistically triggered
	based on the outcomes of already triggered arms.
The reward of the super arm depends on the outcomes of all played arms,
	and it only needs
	to satisfy two mild assumptions, which allow a large class of nonlinear reward
	instances. 	
We assume the availability of an offline $(\alpha,\beta)$-approximation
	oracle that takes the
	means of the outcome distributions of arms and outputs a super arm that
	with probability $\beta$ generates  an $\alpha$ fraction
	of the optimal expected reward.
The objective of an online learning algorithm for CMAB is to minimize {\em $(\alpha,\beta)$-approximation
	regret}, which is the difference in total expected reward
	between the $\alpha\beta$ fraction of expected reward when always
	playing the optimal super arm, and the expected reward of playing
	super arms according to the algorithm.
We provide CUCB algorithm that achieves $O(\log n)$ distribution-dependent regret,
	where $n$ is the number of rounds played, and we further provide distribution-independent
	bounds for a large class of reward functions.
Our regret analysis is tight in that it matches the bound of UCB1 algorithm (up to a constant factor)
	for the classical MAB problem, and it significantly improves the regret bound
	in an earlier paper on combinatorial bandits with linear rewards.
We apply our CMAB framework to two new applications, 	
	probabilistic maximum coverage (PMC) for online advertising and social
	influence maximization for viral marketing, both having nonlinear reward structures.
In particular, application to social influence maximization requires our extension on
	probabilistically triggered arms.
\end{abstract}

\paragraph{keywords:}
combinatorial multi-armed bandit, online learning, upper confidence bound, social influence maximization,
  online advertising.
 

\section{Introduction}

Multi-armed bandit (MAB) is a  problem extensively studied
	in statistics and machine learning.
The classical version of the
	 problem is formulated as a system of $m$ arms (or machines), each
	having an unknown distribution of the reward with an unknown mean.
The task is to repeatedly play these arms in multiple rounds so that the
	total expected reward is as close to the reward of the optimal arm
	as possible.
An MAB algorithm needs to decide which arm to play in the next round given the
	outcomes of the arms played in the previous rounds.
The metric for measuring the effectiveness of an MAB algorithm
	is its {\em regret}, which is the difference in the total expected
	reward between always playing the optimal arm (the arm with the largest expected reward) and playing arms according to
	the algorithm.
The MAB problem and its solutions reflect the fundamental tradeoff
	between exploration and exploitation: 
	whether one should try some arms that have not been played much
	(exploration) or one should stick to the arms that provide good
	reward so far (exploitation).
Existing results show that one can achieve a regret of $O(\log n)$ when
	playing arms in $n$ rounds, and this is asymptotically the best possible.

In many real-world applications, the setting is not the simple MAB one,
	but has a combinatorial nature among multiple arms and possibly non-linear
	reward functions.
For example, consider the following online advertising scenario.
A web site contains a set of web pages and has a set of users visiting the
	web site.
An advertiser wants to place an advertisement on a set of selected web pages
	on the site, and due to his budget constraint, he can select at
	most $k$ web pages.
Each user visits a certain set of pages,
	and each visited page has one click-through
	probability for each user clicking the advertisement
	on the page, but the advertiser does not know these probabilities.
The advertiser wants to repeatedly select sets of $k$ web pages, observe
	the click-through data collected to
	learn the click-through probabilities, and maximize the number of users
	clicking his advertisement over time.

There are several new challenges raised by the above example.
First, page-user pairs can be viewed as arms, but they are not played in isolation.
Instead, these arms form certain combinatorial structures, namely bipartite graphs, 
	and in each round, a set of arms (called a {\em super arm})
	are played together.
Second, the reward structure is not a simple linear function of the outcomes of all
	played arms but takes a more complicated form.
In the above example, for all page-user pairs with the same
	user, the collective reward of these arms
	is either $1$ if the user clicks the advertisement
	on at least one of the
	pages, or $0$ if the user does not click the advertisement on any page.
Third, even the offline optimization problem when the probabilities on all edges of the bipartite graph are known
	is still an NP-hard problem.
Thus, the online learning algorithm needs to deal with combinatorial arm structures, nonlinear reward functions, and 
	computational hardness of the offline optimization task.	
	
Consider another example of viral marketing in online social networks.
In an online social network such as Facebook, companies carry out viral marketing campaigns by 
	engaging with a certain set of seed users (e.g. providing free sample products to seed users), and
	hoping that these seed users could generate a cascade in the network promoting their products.
The cascades follow certain stochastic diffusion model such as the 
	independent cascade model \citep{kempe03}, but the influence probabilities on edges
	are not known in advance and have to be learned over time.
Thus, the online learning task is to repeatedly select seed nodes in a social network, observe
	the cascading behavior of the viral information to learn influence
	probabilities between individuals in the social network, with the goal
	of maximizing the overall effectiveness of all viral cascades.
Similar to the online advertising example given above, we can treat each edge in the social network as a base arm, and
	all outgoing edges from a seed set as a super arm, which is the unit of play.
Besides sharing the same challenges such as the combinatorial arm structures, nonlinear
	reward functions, and computational hardness of the offline maximization task, this viral marketing
	task faces another challenge: in each round after some seed set is selected, the cascade from the seed set may
	probabilistically trigger more edges (or arms) in the network, and the reward of the cascade depends on all probabilistically
	or deterministically triggered arms.

A naive way to tackle both examples above is to treat every super arm as an arm and simply apply the
	classical MAB framework to solve the above combinatorial problems.
However, such naive treatment has two issues.
First, the number of super arms may be exponential to the problem instance
	size due to the combinatorial explosion, and thus classical MAB algorithms
	may need exponential number of steps just to go through all
	the super arms once.
Second, after one super arm is played, in many cases,
	we can observe some information regarding the outcomes
	of the underlying arms, which may  be shared by other
	super arms.
However, this information
is discarded in the classical MAB framework,
	 making it less effective.

In this paper, we define a general framework for the {\em combinatorial
	multi-armed bandit (CMAB)} problem
	to address the above issues 
	and cover a large class of combinatorial online learning
	problems in the stochastic setting, including the two examples given above.
In the CMAB framework, we have a set of $m$ base arms, whose outcomes follow certain unknown
	joint distribution.
A super arm $S$ is a subset of base arms.
In each round, one super arm is played and all base arms contained in the super arm are played.
To accommodate applications such as viral marketing,
	we allow that the play of a super arm $S$ may further trigger more base arms probabilistically, 
	and the triggering depends on the outcomes of the already played base arms in the current round.
The reward of the round is determined by the outcomes of all triggered arms, which are observed as the feedback
	to the online learning algorithm.
A CMAB algorithm needs to use these feedback
	information from the past rounds to decide the super arm to play in the next round.
	
The framework allows an arbitrary combination of arms into super arms.
The reward function only needs to satisfy two mild assumptions 
	\chgins{(referred to as monotonicity and bounded smoothness)}, and thus covering
	 a large class of nonlinear reward functions.
We do not assume the direct
	knowledge on how super arms are formed from underlying
	arms or how the reward is computed.
Instead, we assume the availability of an offline computation oracle that takes
	such knowledge as well as the expectations of outcomes 
	of all arms  as input and
	computes the optimal super arm with respect to the input.

Since many combinatorial problems are computationally hard,
	we further allow (randomized) approximation oracles with failure probabilities.
In particular, we
	relax the oracle to be an $(\alpha,\beta)$-approximation oracle for some
	$\alpha,\beta \le 1$, that is, with success
	probability $\beta$, the oracle could output a super arm whose
	expected reward is at least $\alpha$ fraction of the
	optimal expected reward.
As a result, our regret metric is not comparing against the expected
	reward of playing the optimal
	super arm each time, but against the $\alpha\beta$ fraction of the
	optimal expected reward, since the offline oracle can only guarantee this
	fraction in expectation.
We refer to this as the {\em $(\alpha,\beta)$-approximation regret}.

For the general framework, we provide the
	CUCB (combinatorial upper confidence bound)
	algorithm, an extension to the UCB1 algorithm for the classical MAB
	problem~\citep{AuerCF02}.
We provide a rigorous analysis on the distribution-dependent regret of CUCB and show that
	it is still bounded by $O(\log n)$.
Our analysis further allows us to provide a distribution-independent regret bound that
	works for arbitrary distributions of underlying arms,
	for a large class of CMAB instances.
For the extension accommodating probabilistically triggered arms, 
	we also provide distribution-dependent and -independent bounds
	with triggering probabilities as parameters.

We then apply our framework and provide solutions to two
	new bandit applications, the probabilistic maximum coverage
	problem for advertisement placement and social influence maximization
	for viral marketing.
The offline versions of both problems are NP-hard, with constant approximation algorithms
	available.
Both problems have nonlinear reward structures that cannot be handled by any existing work.

We also apply our result to combinatorial bandits with linear rewards,
	recently studied by~\cite{Yi2012}.
We show that we significantly improve their distribution-dependent regret bound, even though we are covering a much
	larger class of combinatorial bandit instances.
We also provide new distribution-independent bound not available in~\citep{Yi2012}.

This paper is an extension to our ICML'13 paper \citep{CWY13}, with explicit modeling of probabilistically triggered arms and
	their regret analysis for the CUCB algorithm.
We correct an erroneous claim in \citep{CWY13}, which states that the original CMAB model and result without probabilistically
	triggered arms can be applied to the online learning task for social influence maximization.
Our correction includes explicit modeling of probabilistically triggered arms in the CMAB framework, and significant reworking of
	the regret analysis to incorporate triggering probabilities in the analysis and the regret bounds.

In summary, our contributions include:
	(a)  defining a general CMAB framework
	that encompasses a large class of nonlinear
	reward functions,
	(b) providing CUCB algorithm with a rigorous regret analysis as a general solution to this
	CMAB framework,
	(c) further generalizing our framework to accommodate probabilistically triggered base arms, and 
	applying this framework to the social influence maximization 
	problem, and 
	(d) demonstrating that our general framework can be
	effectively applied to a number of practical combinatorial bandit problems, including ones
	with nonlinear rewards.
Moreover, our framework provides a clean separation of the online learning task
	and the \mbox{offline} computation task: the oracle takes care of the offline computation task,
	which uses the domain knowledge of the problem instance, while
	our CMAB algorithm takes care of the online learning task, and is oblivious
	to the domain knowledge of the problem instance.

\paragraph{Related work.}
Multi-armed bandit problem has been well studied in the literature,
	in particular in statistics and reinforcement learning \citep[cf.][]{BF85,SB98}.
Our work follows the line of research on stochastic MAB problems,
 	which is initiated by \cite{LaiRobbins85}, who show that
	under certain conditions on reward distributions, one can achieve
	a tight asymptotic regret of $\Theta(\log n)$, where $n$ is
	the number of rounds played.
Later, \cite{AuerCF02} demonstrate that $O(\log n)$ regret can be achieved
	uniformly over time rather than only asymptotically.
They propose several MAB algorithms, including the UCB1 algorithm, which has been
	widely followed and adapted in MAB research.

For combinatorial multi-armed bandits, a few specific instances
	of the problem has been studied in the literature.
A number of studies consider simultaneous plays of
	$k$ arms among $m$ arms \citep[e.g.][]{AVW87a,CG07,LiuLZ11}.
Other instances include the matching
bandit~\citep{GaiKJ10} and the online shortest path problem~\citep{Liu2012}.

The work closest to ours is a recent work by \cite{Yi2012},
	which also considers a combinatorial bandit framework with an approximation
	oracle.
However, our work differs from theirs in several important aspects.
Most importantly, their work only considers linear rewards while our CMAB framework includes
	a much larger class of linear and nonlinear rewards.
Secondly, our regret analysis is much tighter, and as a result we significantly improve
	their distribution-dependent regret bound when applying our result to the linear reward case, and we are able
	to derive a distribution-independent regret bound close to the theoretical lower bound
	while they do not provide distribution-independent bounds.
Moreover, we allow the approximation oracle to have a failure probability (i.e., $\beta < 1$),
	while they do not consider such failure probabilities.

In terms of types of feedbacks in combinatorial bandits~\citep{Audibert2011},
	our work belongs to the
	{\em semi-bandit} type, in which the player observes only the outcomes of played arms
	in one round of play.
Other types include (a) {\em full information}, in which the player observes the outcomes of 	
	all arms, and (b) {\em bandit}, in which the player only observes the final
	reward but no outcome of any individual arm.
More complicated feedback dependences are also considered by~\cite{Mannor2011}.

\chgins{
Bounded smoothness property in our paper is an extended form of Lipschitz condition, but our model and results differ from
	the Lipschitz bandit research \citep{Bobby1} in several aspects.
First, Lipschitz  bandit considers a continuous metric space where every point is an arm, and the Lipschitz condition is applied to 
	two points (i.e., two arms). 
Under this assumption, if we know one arm pretty well, we will also know the nearby arms pretty well. In contrast,
our bounded smoothness condition is applied to a vector of mean values of the base arms instead of one super arm, and by knowing one super arm well, 
	we cannot directly know how good are the other super arms. 
Second, the feedback model is different: Lipschitz bandit assumes bandit feedback model
	while our CMAB assumes semi-bandit feedback.
Third, using the Lipschitz condition, 
they designed a new algorithm called zooming algorithm, which maintains a confidence radius of arms, so that by knowing the center arm well, they are also pretty confident of the arms within the confidence radius of the center. 
In comparison, our algorithm is basically a direct extension of the classical UCB algorithm, in which the confidence radius is used to get confidence on the estimate
	of every base arm.}
\chgins{
\cite{Bobby1} generalize the setting of continuum bandits, which assumes the strategy set is a compact subset of 
$\mathbb{R}^d$ and the reward function satisfies the Lipschitz condition, see e.g. \citep{continuum2, continuum1}.
}


A different line of research considers {\em adversarial multi-armed bandit},
	initiated by \cite{AuerCFS02},
	\chgins{ in which no probabilistic assumptions are made about the rewards, and they can even be chosen by an adversary. 
	}
In the context of adversarial bandits, several studies also consider combinatorial
	bandits~\citep{CL09,Audibert2011,Bubeck12}.
For linear rewards, \cite{Kakade2009} have shown how to convert an
	approximation oracle into an online algorithm with sublinear regret both in the full information setting and the bandit setting.
For non-linear rewards, various online submodular optimization problems with bandit feedback are studied
	in the adversarial setting~\citep{Streeter2008, Radlinski2008, Streeter2009, Hazan2009}.
Notice that our framework deals with stochastic instances and we can handle reward functions more general than
the submodular ones.

This paper is the full version of our ICML'13 paper \citep{CWY13} with the extension to include probabilistically triggered arms
	in the model and analysis.
We made a mistake in \citep{CWY13} by claiming that the online learning task for social influence maximization is an instance
	of the original CMAB model proposed in \citep{CWY13} without explicitly modeling probabilistically triggered arms.
In this paper we correct this mistake by allowing probabilistically triggered arms in the CMAB model, and by significantly
	revising the analysis to include triggering probabilities in the analysis and the regret bounds.

Since our work in \citep{CWY13}, several studies are also related to combinatorial multi-armed bandits or in general combinatorial online
	learning.
\cite{QinCZ14} extend CMAB to contextual bandits and apply it to diversified online recommendations.
\cite{LAKLC14} address combinatorial actions with limited feedbacks.
\cite{GMM14} use Thompson sampling method to tackle combinatorial online learning problems. 
Comparing with our CMAB framework, they allow more feedback models than our semi-bandit feedback model, but
	they require finite number of actions and observations, their regret contains a large constant term, and it is unclear if their framework supports
	approximation oracles for hard combinatorial optimization problems.
\cite{KWAEE14} study linear matroid bandits, which is a subclass of the linear combinatorial bandits we discussed in Section~\ref{sec:linear}, and
	they provide better regret bounds than our general bounds given
	in Section~\ref{sec:linear}, because their analysis utilizes the matroid combinatorial structure.
\chgins{In a latest paper \cite{KWAS15} improve the regret bounds of the linear combinatorial bandits
	via a more sophisticated non-uniform sufficient sampling condition than the one we used in
	our paper.
However, it is unclear if this technique can be applied to non-linear reward functions satisfying
	the bounded smoothness condition (see discussions in Section~\ref{sec:linear} for more
	details).}

\paragraph{Paper organization.}
In Section~\ref{sec:model} we formally define the CMAB framework.
Section~\ref{sec:alg} provides the CUCB algorithm and the main results on its regret bounds and the proofs.
Section~\ref{sec:app} shows how to apply the CMAB framework and CUCB algorithm to the online advertising and
	viral marketing applications, as well as the class of combinatorial bandits with linear reward functions.
We conclude the paper in Section~\ref{sec:conclude}.

\vspace{\sectionspace}
\section{General CMAB Framework}
\label{sec:model}
A {\em combinatorial multi-armed bandit (CMAB)} problem consists of $m$ {\em base arms}
	associated with a set of random variables
$X_{i,t}$ for $1\leq i\leq m$ 
and $t\geq 1$, 
with bounded support 
on $[0,1]$. 
Variable $X_{i,t}$ indicates the random outcome of the $i$-th base arm in its
	$t$-th trial.
The set of random
variables $\{X_{i,t} \mid t \geq 1\}$  associated with base arm $i$ are
independent and identically distributed according to some unknown
distribution with unknown expectation $\mu_i$.
Let $\bp = (\mu_1,\mu_2,\ldots, \mu_m)$ be the vector of expectations of
	all base arms.
Random variables of different base arms may be dependent.

The unit of play in CMAB is a {\em super arm}, which is a set of base arms.
Let $\cal S$ denote the set of all possible super arms that can be played in a CMAB problem instance.
For example, $\cal S$ could be the set of all subsets of base arms containing at most $k$ base arms.
In each round, one of the super arms $S \in {\cal S}$ is selected and played, and every
	base arm $i\in S$ are triggered and played as a result.
The outcomes of base arms in $S$ may trigger other base arms not in $S$ to be played, and the outcomes
	of these arms may further trigger more arms to be played, and so on.
Therefore, when super arm $S$ is played in round $t$, a superset of $S$ is triggered and played, and
	the final reward of this round depends on the outcomes of all triggered base arms.
The feedback in the round after playing super arm $S$
	is the outcomes of the triggered (played) base arms.
The random outcomes of triggered base arms in one round are independent of random outcomes in other
	rounds, but they may depend on one another in the same round.

For each $i\in [m]$, let $p_i^S$ denote the probability that base arm $i$ is triggered when super arm
	$S$ is played.
Once super arm $S$ is fixed, the event of triggering of base arm $i$ is independent of the history of
	previous plays of super arms.
It is clear that for all $i\in S$, $p_i^S = 1$.
Note that probability $p_i^S$ may not be known to the learning algorithm, since the event of triggering base 
	arm $i$ may depend on the random outcomes of other base arms, the distribution of which may be unknown.
Moreover, the triggering of base arms may depend on certain combinatorial structure of
	the problem instance, and
	triggering of different base arms may not be independent from one another
	(for an example, see the social influence maximization application in Section~\ref{sec:infmax}).
	
Let $\trig{S} = \{i \in [m]\,|\, p_i^S > 0 \}$ denote the set of possibly triggered base 
	arms by super arm $S$, also referred to as the {\em triggering set} of $S$.
Let $p_i\triangleq \min_{S\in {\cal S}, i\in \trig{S}} p_i^{S}$ denote the minimum nonzero 
	triggering probability of base arm $i$ under all super arms.
When $p_i=1$ for all $i\in [m]$, each super arm $S$ deterministically triggers all base arms in $\trig{S}$,
	in which case we treat $S$ and $\trig{S}$ as the same set.
\chgred{Let $p^* \triangleq \min_{i\in [m]} p_i$.}

\chgins{
In our model, it is possible that a base arm $i$ does not belong to any super arm, and thus $i$
	can only be probabilistically triggered.
In fact, our model is flexible enough to allow that {\em all} based arms are probabilistically triggered.
To do so, we can simply add a set of dummy base arms and dummy super arms 
	for the purpose of probabilistically triggering real base arms.
In particular, for each real base arm $i$, we can add a dummy base arm $d_i$, which is
	a Bernoulli random variable with $1$ meaning $i$ is triggered and $0$ meaning $i$ is not
	triggered.
Then a dummy super arm containing a subset of these Bernoulli dummy base arms can be used to
	probabilistically trigger a set of real base arms.
If all super arms are such dummy super arms, then all real base arms are only probabilistically
	triggered.
}


For each arm $i\in [m]$, let 
         $T_{i,t}$ denote the number of times
arm $i$ has been successfully triggered after the first $t$ rounds in which
	$t$ super arms are played.
If an arm $i \in \trig{S}\setminus S$ is not triggered in round $t$ when super arm $S$ is played, 
	then $T_{i,t} = T_{i,t-1}$.
Let $R_t(S)$ be a non-negative
	random variable denoting the reward of round $t$ when
	super arm $S$ is played.
The reward depends on the actual problem instance definition, the
	super arm $S$ played, and the outcomes of all triggered arms in
	round $t$.
The reward $R_t(S)$  might be as simple as a summation of the outcomes
	of the triggered arms in $S$:
	$R_t(S) = \sum_{i\in \trig{S}, i \mathrm{~is~triggered}}X_{i,T_{i,t}}$,
	but our framework allows more sophisticated nonlinear rewards, as explained below.

In this paper, we consider CMAB problems in which the expected reward
	of playing any super arm $S$ in any round $t$,
	$\E[R_t(S)]$, is a function of $S$ and the expectation vector
	$\bp$ of all arms.
For the linear reward case as given above together with no probabilistic triggering
	($S=\trig{S}$), this is true because linear
	addition is commutative with the expectation operator.
For non-linear reward functions not commutative with the expectation operator, it is still true
	if we know the type of distributions and only the expectations of arm outcomes are unknown.
For example, the distribution of $X_{i,t}$'s are known to be independent $0$-$1$ Bernoulli random variables
	with unknown mean $\mu_i$.\chgins{\footnote{\chgins{It is also possible that the Bernoulli
	random variables are not independent. For example, the joint distribution is determined
	by sampling a random value $\rho\in [0,1]$ uniformly at random, and then each base arm $i$
	takes value $1$ if any only if $\rho \le \mu_i$.}}}
Henceforth, we denote the expected reward of playing $S$ as
	$r_\bp(S)\triangleq \E[R_t(S)]$.
\begin{mydef}[Assumptions on expected reward function] \label{def:assumptions}
To carry out our analysis, we make the following two mild assumptions
	on the expected reward $r_\bp(S)$:
\begin{itemize} 
\item {\bf Monotonicity}. The expected reward of playing any
	super arm $S\in \calS$  is
  monotonically nondecreasing with respect to the expectation vector,
	i.e., if for all $i\in [m]$, $\mu_i \leq \mu_i' $, we have
	$r_{\bp}(S) \leq r_{\bp'}(S)$ for all $S\in \calS$.
\item {\bf Bounded smoothness.}
There exists a continuous, strictly increasing (and thus invertible) function
	$f(\cdot)$ with $f(0)=0$, called {\em bounded smoothness function}, such that
	for any two expectation vectors $\bp$ and $\bp'$ and for any $\Lambda > 0$,
	we have $|r_{\bp}(S) - r_{\bp'}(S)| \leq f(\Lambda)$ if
	$\max_{i\in \trig{S}}|\mu_i - \mu_i'|\le \Lambda$. 
\end{itemize}
\end{mydef}
Both assumptions are natural. In particular, they hold true for
  all the applications we considered.
\chgins{We remark that bounded smoothness is an extended form of Lipschitz condition in that we use a general function $f$ instead of linear or
	power-law functions typically used in Lipschitz condition definition, and we use infinity norm instead of
	typically used $L_2$ norm.}

\begin{mydef}[CMAB algorithm]
A CMAB algorithm $A$ is one that selects the super arm of
	round $t$ to play based on the outcomes of revealed arms of previous
	rounds, without knowing the expectation vector $\bp$.
Let $S^A_t \in {\cal S}$ be the super arm selected by $A$ in round $t$.
Note that $S^A_t$ is a random super arm that depends on the outcomes
	of arms in previous rounds and potential randomness in the algorithm
	$A$ itself.
The objective of algorithm $A$
	is to maximize the expected reward of all rounds up to
	a round $n$, that is,
	$\E_{S,R}[\sum_{t=1}^n R_t(S^A_t)]=\E_S[\sum_{t=1}^n r_\bp(S^A_t)]$,
	where $\E_{S,R}$ denotes taking expectation among all random events
	generating the super arms $S^A_t$'s and generating rewards $R_t(S^A_t)$'s,
	and $\E_S$ denotes taking expectation only among all random events
	generating the super arms $S^A_t$'s.
\end{mydef}

We do not assume that the learning algorithm has the direct knowledge about the
	problem instance, e.g. how super arms are formed from the base
	arms, how base arms outside of a super arm are triggered, and how reward is defined.
Instead, the algorithm has access to a computation oracle that
	takes the expectation vector $\bp$ as the input, and together with
	the knowledge of the problem instance, computes the optimal or near-optimal
	super arm $S$.
Let $\opt_\bp = \max_{S\in \calS}r_\bp(S)$ and
	$S_\bp^* = \argmax_{S\in \calS}r_\bp(S)$.
We consider the case that exact computation of
	$S_\bp^*$ may be computationally hard, and the algorithm may be
	randomized with a small failure probability.
Thus, we resolve to the following {\em $(\alpha,\beta)$-approximation oracle}:
\begin{mydef}[$(\alpha,\beta)$-Approximation oracle] 
For some $\alpha,\beta \le 1$,
	$(\alpha,\beta)$-approximation oracle is an oracle
	that takes an expectation vector
	$\bp$ as input, and outputs a super arm $S\in \calS$, such that
	$\Pr[r_{\bp}(S) \geq {\alpha} \cdot \opt_{\bp}]\geq \beta$.
    Here $\beta$ is the success probability of the oracle.
\end{mydef}

Many computationally hard problems do admit efficient approximation oracles~\citep{Vazirani04}.
With an $(\alpha,\beta)$-approximation oracle, it is no longer fair
	to compare the performance of a CMAB algorithm against the optimal
	reward $\opt_\bp$ as the regret of the algorithm.
Instead, we compare against the $\alpha\cdot\beta$ fraction of the
	optimal reward, because only a $\beta$ fraction of oracle computations
	are successful, and when successful the reward is only an $\alpha$-approximation
	of the optimal value.
\begin{mydef}[$(\alpha,\beta)$-approximation regret]	\label{def:regretdef}
$(\alpha,\beta)$-approximation regret of
	a CMAB algorithm $A$ after $n$ rounds of play
	 using an $(\alpha,\beta)$-approximation oracle under the expectation vector $\bp$
is defined as
\begin{equation}
Reg^A_{\bp,\alpha,\beta}(n) =
	n \cdot \alpha  \cdot \beta\cdot  \opt_{\bp}  -
	\E_S\left[ \sum_{t=1}^n r_\bp(S^A_t)\right].
\end{equation}
\end{mydef}
Note that the classical MAB problem is a special case of our
	general CMAB problem, in which
	(a) the constraint $\calS = \{\{i\}\,|\, i\in [m]\}$ so that
	each super arm is just a base arm;
	(b) $S = \trig{S}$ for all super arm $S$, that is, playing of a base arm does not trigger any
	other arms; 
	(c) the reward of a super arm $S=\{i\}$ in its
	$t$'s trial is its outcome $X_{i,t}$;
	(d) the monotonicity and bounded smoothness hold trivially with
	function $f(\cdot)$ being the identity function; and
	(e) the $(\alpha,\beta)$-approximation oracle is simply the $\argmax$ function
	among all expectation vectors, with $\alpha=\beta=1$.

\vspace{\sectionspace}
\section{CUCB Algorithm for CMAB}
\label{sec:alg}

\begin{algorithm}[t]
    \centering
    \caption{CUCB with computation oracle.
\label{alg:oracleucb}
}
    \label{algorithmoracle}
    \begin{algorithmic}[1]
        \STATE For each arm $i$, maintain: (1) variable $T_{i}$
		as the total number of
        times arm $i$ is played so far\chgins{, initially $0$};
	(2) variable $\hat \mu_{i}$ as the
	mean of all outcomes $X_{i,*}$'s of arm $i$ observed so far\chgins{, initially $1$}.
\chgins{        \STATE $t \leftarrow 0$.}
        \WHILE {\TRUE}
            \STATE $t \leftarrow t+1$.
            \STATE For each arm $i$,
                  \chgins{set $\bar{\mu}_{i}=\min\left \{\hat{\mu}_{i} + \sqrt{\frac{
	3\ln t}{2T_{i}}},1\right \}$}. \label{alg:adjustCUCB}
            \STATE \label{alg:selectsarm} $S = \mathrm{Oracle}(\bar{\mu}_{1},
            \bar{\mu}_{2},\ldots, \bar{\mu}_{m})$.
            \STATE Play $S$, observe outcomes of played base arms $i$,
             and update all $T_{i}$'s and $\hat \mu_{i}$'s.
        \ENDWHILE
    \end{algorithmic}
\end{algorithm}

We present our CUCB algorithm
in Algorithm~\ref{alg:oracleucb}.
We maintain an empirical mean $\hat \mu_i$ for each arm
	$i$. More precisely, if arm $i$ has been played $s$ times
	by the end of round $n$, then the value of
	$\hat \mu_i$ at the end of round $n$ is $(\sum_{j=1}^{s}X_{i,j})/s$.
The actual expectation vector $\bar{\bp}$ given to the
oracle contains an adjustment term $\sqrt{\frac{3\ln t}{2T_{i}}}$
	 for each $\hat \mu_i$,
	which depends on the round number $t$ and the number of times
	arm $i$ has been played (stored in variable $T_{i}$).
Then we simply play
the super arm returned by the oracle and update variables
	$T_{i}$'s and $\hat \mu_i$'s	
	accordingly.
Note that in our model all arms have bounded support on $[0,1]$, but
	with the adjustment \chgins{the upper confidence bound 
	$\hat{\mu}_{i} + \sqrt{\frac{3\ln  t}{2T_{i}}}$ may exceed $1$, in which
	case we simply trim it down to $1$ and assign it to $\bar{\mu}_i$ (line~\ref{alg:adjustCUCB}).}

\chgins{
Our algorithm does not have an initialization phase where all base arms are played at least once.
This is to accommodate the case where some base arms may only be probabilistically triggered
	and there is no super arm that can trigger them deterministically.
Instead, we simply initialize the counter $T_i$ to $0$ and $\hat{\mu}_i$ to $1$
	for every base arm $i$.
Thus initially $\bar{\mu}_i=1$ for all $i$, and the oracle will select a super arm given an all-one
	vector input.
Intuitively, any base arm $i$ that has not been played will have its $\bar{\mu}_i=1$, which should
	let the oracle be biased toward playing a super arm that (likely) triggers $i$.
It may be possible that a base arm $i$ is never played, and this only means that $i$ is not important
	for the optimization task and the oracle decides not to play it (deterministically or probabilistically).
Our analysis works correctly without the initialization phase.
}

We now provide necessary definitions for the main theorems.
\begin{mydef}[Bad super arm] \label{def:badsuperarm}
A super arm $S$ is {\em bad} if $r_\bp(S) < \alpha \cdot \opt_\bp$. 
The set of bad super arms
is defined as
$\calS_{\bad} \triangleq \{ S\,\mid \, r_{\bp}(S) < \alpha \cdot \opt_{\bp}\}$. 
For a given base arm $i\in [m]$, 
	let $\calS_{i,\bad} = \{ S \in \calS_{\bad} \,|\, i \in \trig{S}  \}$ be
	the set of bad super arms whose triggering sets contain $i$.
We sort all bad super arms in $\calS_{i,\bad}$ as $S_{i,\bad}^1, S_{i,\bad}^2, \ldots, S_{i,\bad}^{K_i}$,
 in increasing order of their expected rewards, where $K_i = |\calS_{i,\bad}|$.
Note that when $K_i=0$, there is no bad super arm that can trigger base arm $i$.	
\end{mydef}

\begin{mydef}[$\Delta$ of bad super arms]\label{def:delta}
For a bad super arm $S \in \calS_{\bad} $, we define 
	$\Delta_S \triangleq \alpha \cdot \opt_{\bp} - r_{\bp}(S)$.
For a given base arm $i\in [m]$ with $K_i>0$ and index $j \in [K_i]$, we define
\begin{equation*}
\label{eqn:deltaij}
\Delta^{i,j} \triangleq \Delta_{S_{i,\bad}^j}.
\end{equation*}
We have special notations for
the minimum and the maximum $\Delta^{i,j}$
for a fixed $i$ with $K_i>0$:
\begin{align*}
\Delta_{\max}^i &\triangleq \Delta^{i,1},\\
\Delta_{\min}^i &\triangleq \Delta^{i, K_i}.
\end{align*}
Furthermore, define $\Delta_{\max}\triangleq \max_{i\in [m], K_i>0} \Delta_{\max}^i$,
	$\Delta_{\min}\triangleq\min_{i\in [m], K_i>0} \Delta_{\min}^i$.

\end{mydef}

Our main theorem below provides the distribution-dependent regret bound of  the CUCB algorithm using the $\Delta$ notations.
We use $\I\{\cdot\}$ to denote the indicator function, and $\I \{{\cal E}\} 
= 1$ if
	$\cal E$ is true, and $0$ if $\cal E$ is false.
\begin{mythm}
\label{thm:cucb individual}
	The $(\alpha,\beta)$-approximation regret of the CUCB algorithm in $n$ rounds
	using an $(\alpha,\beta)$-approximation
	oracle is at most
\begin{align}
&\sum_{i\in [m], K_i>0} \left(
\ell_n(\Delta^i_{\min},p_i) \Delta^i_{\min} + \int_{\Delta^i_{\min}}^{\Delta^i_{\max}} \ell_n(x,p_i) \mathrm{d}x
\right ) + \chgred{\left(\frac{(2+\I\{p^* < 1\})\pi^2}{6} +1\right)\cdot m \cdot  \Delta_{\max}}, \label{eqn:detailed regret bound}
\end{align}
where \chgred{$p^* = \min_{i\in [m]} p_i$, and }
\chgred{
\begin{equation*}
\ell_n(\Delta,p)=
\left\{
\begin{array}{lr}
\max\left (\frac{12\cdot \ln  n}{(f^{-1}(\Delta))^2\cdot p}, 
\frac{24\cdot \ln  n}{p} \right ), & \textrm{if~} 0<p<1,\\
\frac{6\ln  n}{(f^{-1}(\Delta))^2},& \textrm{if~} p=1.
\end{array}
\right.
\end{equation*}}
and $f(\cdot)$ is the bounded smoothness function.
\end{mythm}

\chgred{Note that when $p_i=1$ for all $i\in [m]$, each super arm $S$ deterministically triggers base arms
	in $S$ and no probabilistic triggering of other arms.
In this case, the above theorem falls back to Theorem 1 of \cite{CWY13}.
When $p_i<1$ for some $i\in [m]$, the regret bound is slightly more complicated, in particular, it has
	an extra factor of $1/p_i$ appearing in the leading $\ln n$ term.}

In Theorem~\ref{thm:cucb individual}, when $\Delta^i_{\min}$ is extremely small, the regret would be approaching infinity. 
Below we prove
a distribution-independent regret for arbitrary
	distributions with support in $[0,1]$ on all arms, for a large class of problem instances
	with a polynomial bounded smoothness function $f(x) = \gamma x^\omega$ for $\gamma > 0$ and $ 0< \omega \leq 1$.
The rough idea of the proof is, if $\Delta^i_{\min}\leq 1/\sqrt n$, it can only contribute $\sqrt n$ regret at time horizon $n$.
The proof of the following theorem relies on the tight regret bound of Theorem~\ref{thm:cucb individual} on the leading $\ln n$ term.

\def\thmworst{
Consider a CMAB problem with an $(\alpha,\beta)$-approximation oracle. 
Let $p^* = \min_{i\in [m]} p_i$.
If the bounded smoothness function $f(x) = \gamma \cdot x^\omega$ for some $\gamma >0$ and $\omega\in (0,1]$, the regret of
	CUCB
	is at most:
\[ 
\left\{
\begin{array}{lr}
\frac{2\gamma}{2-\omega}\cdot (6m\ln  n)^{\omega/2} \cdot n^{1-\omega/2}+
\left(\frac{\pi^2}{3}+1\right)\cdot m \cdot \Delta_{\max},  & \textrm{if~} p^*=1,\\
\frac{2\gamma}{2-\omega}\cdot \left(\frac{12m\ln  n}{p^*}\right)^{\omega/2} \cdot n^{1-\omega/2}+
\left(\chgred{\frac{\pi^2}{2}}+1\right)\cdot m \cdot \Delta_{\max} + \sum_{i\in [m]} \chgred{\frac{24\ln  n}{p_i} }  \cdot \Delta_{\max}, & \textrm{if~} 0<p^*<1.\\
\end{array}
\right.
\]}

\begin{mythm} \label{thm:worstbound}
{\thmworst}
\end{mythm}

Note that for all applications discussed in Section~\ref{sec:app}, we have $\omega=1$.
For the classical MAB setting with $\omega=1$ and $p^* = 1$, 
	we obtain a distribution-independent bound of $O(\sqrt{mn\ln n})$, which matches
	(up to a logarithmic factor)	the original UCB1 algorithm~\citep{Audibert2009}.
In the linear combinatorial bandit setting, i.e., semi-bandit with $L_\infty$ assumption in~\cite{Audibert2011}, our regret is $O(\sqrt{m^3n\log n})$, which is a factor $\sqrt{m}$ off the optimal bound in the adversarial setting, \chgins{a more general setting than
	the stochastic setting (see the discussion in the end of
	Section~\ref{sec:linear} for a reason of this gap)}.

\subsection{Proof of the Theorems}

\subsubsection{Proof of Theorem~\ref{thm:cucb individual}}

Before getting to the proof of our theorem, 
we need more definitions and lemmas. First, we have a convenient 
notation for the case when the oracle outputs 
non-$\alpha$-approximation answers.

\begin{mydef}[Non-$\alpha$-approximation output] \label{def:nonalphaoutput}
In the $t$-th round, let $F_t$ be the event
that the oracle fails to produce an $\alpha$-approximate
answer with respect to its input $\bar{\bp}=(\bar{\mu}_{1},
            \bar{\mu}_{2},\ldots, \bar{\mu}_{m})$. We have
$\Pr[F_t]=\mathbb{E}[\I\{F_t\}]\leq 1-\beta$.
\end{mydef}

Since the value of many variables are changing 
in different rounds, we also define notations for their value in round $t$. All of them are random variables.

\begin{mydef}[Variables in round $t$]
For variable $T_i$, let $T_{i,t}$ be the value of $T_i$ at the end of round $t$,
	that is, $T_{i,t}$ is the number of times arm $i$ is played in the
	first $t$ rounds.
For variable $\hat \mu_i$, let $\hat \mu_{i,s}$ be the value of $\hat \mu_i$ after
	arm $i$ is played $s$ times, that is,
	$\hat \mu_{i,s} = (\sum_{j=1}^{s} X_{i,j})/s$, where $X_{i,j}$ is the outcome of base arm $i$ in its $j$-th trial,
	as defined at the beginning of Section~\ref{sec:model}.
Then, the value of variable $\hat \mu_i$ at the end of round $t$
	is $\hat \mu_{i,T_{i,t}}$.
For variable $\bar \mu_i$, let $\bar \mu_{i,t}$ be the value of $\bar \mu_i$
	at the end of round $t$.
\end{mydef}

Next, we introduce an important 
parameter in our proof called sampling threshold.

\chgred{
\begin{mydef}[Sampling threshold] \label{def:samplthreshold}
For a probability value $p\in (0,1]$ and reward difference value $\Delta \in \R^+$, the value
	$\ell_n(\Delta,p)$ defined  below is called the {\em sampling threshold} for round $n$:
\begin{equation*}
\ell_n(\Delta,p)=
\left\{
\begin{array}{lr}
\max\left (\frac{12\cdot \ln  n}{(f^{-1}(\Delta))^2\cdot p}, 
\frac{24\cdot \ln  n}{p} \right ), & \textrm{if~} 0<p<1,\\
\frac{6\ln  n}{(f^{-1}(\Delta))^2},& \textrm{if~} p=1.
\end{array}
\right.
\end{equation*}

\end{mydef}
}

Informally, base arm $i\in [m]$ at round $n$ is considered as sufficiently sampled if the number of
	times $i$ has been played by round $n$, $T_{i,n}$, is above its sampling threshold
	$\ell_n(\Delta^i_{\min},p_i)$.
When all base arms are sufficiently sampled, with high probability we would obtain
	accurate estimates of their sample means and would be able to distinguish the $\alpha$-approximate 
	super arms from bad super arms.


\chgred{
We utilize the following well known tail bounds in our analysis.

\begin{myfact}[Hoeffding's Inequality \citep{hoeffding63}]\label{fact:hoeffding}
Let $X_1, \cdots , X_n$ be independent and identically distributed random variables with
common support $[0,1]$ and mean $\mu$. 
Let
$Y=X_1+\cdots+X_n$. Then for all $\chgins{\delta} \geq0$,
\[\Pr\{|Y-n\mu|\geq \chgins{\delta} \} \leq 2e^{-2\chgins{\delta}^2/n}.
\]
\end{myfact}
%
%

\begin{myfact}[Multiplicative Chernoff Bound \citep{MU05}\footnote{\chgred{The result in \citep{MU05} 
	(Theorem 4.5 together with Excercise 4.7) only covers the case where random variables $X_i$'s are independent. 
However the result can be easily generalized to our case with an almost identical proof.
The only main change is to replace $\E\left[e^{{t(\sum_{j=1}^{i-1} X_j + X_i)}}\right]=
	\E\left[e^{{t\sum_{j=1}^{i-1} X_j}}\right]\E\left[e^{{t X_i}}\right]$
	with $\E\left[e^{t(\sum_{j=1}^{i-1} X_j + X_i)}\right] = 
	\E\left[e^{t\sum_{j=1}^{i-1} X_j} \E\left[e^{tX_i}\mid X_1,\ldots, X_{i-1}\right]\right]$.}}]\label{chernoffLemma}
Let $X_1, \cdots , X_n$ be Bernoulli random variables taking values from $\{0,1\}$, 
	and $\E[X_t|X_1,\cdots, X_{t-1}]\geq\mu$ for every $t\leq n$. 
Let $Y=X_1+\cdots+X_n$. 
Then for all $0<\chgins{\delta} <1$,
\[\Pr\{Y\leq (1-\chgins{\delta})n\mu\} 
\leq e^{-\frac{\chgins{\delta}^2n\mu }{2}}.
\]
\end{myfact}
}



Using the above tail bounds, we can prove that 
with high probability,
the empirical mean of a set of independently sampled variables is close to the actual mean. 
Below we give a definition on the standard difference between 
the empirical mean and the actual expectation.

\begin{mydef}[Standard difference]\label{def:standdif}
For the random variable $T_{i,t-1}$, 
standard difference 
is 
defined as a random variable
\chgins{$\Lambda_{i,t}=\min\{\sqrt{ \frac{
3\ln  t}{2T_{i,t-1}}},1\}$.} 
The maximum standard difference 
is defined as a random variable 
 $\Lambda_{t}
=\max \{
\Lambda_{i,t} \,\mid i\in \trig{S}_t\}$ (be reminded that it is $\trig{S}_t$, not $S_t$).
\chgins{The universal 
difference bound is defined as 
$\Lambda^{i,l} = 
\frac{f^{-1}(\Delta^{i,l})}{2}$, 
which is \emph{not} a random variable.} 
\end{mydef}

If in the round $t$, 
the difference between the empirical mean and the actual 
expectation is below the standard difference, we call the process 
a ``nice process''. See the formal definition below. 
\begin{mydef}[Nice run]\label{def:nicerun}
The run of Algorithm~\ref{alg:oracleucb} is {\em nice} at time $t$ (denoted as the indicator $\calN_t$) if:
\begin{align}
	\forall i\in [m], \, |\hat{\mu}_{i,T_{i,t-1}} -\mu_i| \,\chgins{\le}\, 
	\Lambda_{i,t}.\label{eqn:niceProcess} 
\end{align}
\end{mydef}

\begin{mylem}
\label{lem:nice}
The probability that the run of Algorithm~\ref{alg:oracleucb} is {\em nice} at time $t$ is at least $1-\frac{2m}{ t^2}$.
\end{mylem}
\begin{proof}

\chgins{
If $T_{i,t-1}=0$, this is trivially true. 
If $T_{i,t-1}>0$, 
by} the Hoeffding's inequality in Fact~\ref{fact:hoeffding},
for any $i\in [m]$,
\begin{align}
&\Pr\left\{
\mid \hat \mu_{i,T_{i,t-1}} - \mu_i\mid
\geq  \Lambda_{i,t}
\right\}  = \sum_{s=1}^{t-1} \Pr
\left\{ \mid \hat \mu_{i,s} - \mu_i\mid
\geq  \Lambda_{i,t}, T_{i,t-1} = s
\right\} \nonumber\\
 \leq &\sum_{s=1}^{t-1} \Pr
\left\{ \mid \hat \mu_{i,s} - \mu_i\mid
\geq  \sqrt{\frac{
3\ln  t
}{2s} }
\right\}
\leq t\cdot 2 e^{-3\ln  t}
= \frac{2}{ t^2}. \label{eqn:hatp-p:appendix}
\end{align}
The lemma follows by taking union bound on $i$.
\end{proof}

Lemma \ref{lem:nice} tells us that if at time $t$, 
$T_{i,t-1}$ is large, then 
we can get a good estimation of $\mu_i$. 
Intuitively, if we estimate all $\mu_i$'s pretty well, 
it is unlikely that we will choose a bad 
super arm using the approximation oracle. 
On the other hand, in the case that for some $i$ $T_{i,t-1}$ is small, 
although we may not have a good estimate of $\mu_i$, 
it indicates that arm $i$ has not been played for 
many times, which gives us an upper bound on the number of times 
that the algorithm plays a bad super arm containing arm $i$.  
Based on this idea, it is crucial to find a 
sampling threshold, which separates these two cases.

Now we need to define the way that we count the sampling times of each arm $i$. 

\begin{mydef}[Counter for arm $i$]\label{def:counters}
We maintain a counter $N_i$ 
for each arm $i$.
Let $N_{i,t}$
  be the value of $N_i$ at the end of round $t$ and $N_{i,0} = 0$.
  $\{N_i\}$ is updated in the following way.

\chgins{For a round $t>0$},
 let $S_t$ be the super arm selected in round $t$ by the oracle (line~\ref{alg:selectsarm} of Algorithm~\ref{alg:oracleucb}).
  Round $t$ is {\em bad} if the oracle selects a
  {\em bad} super
  arm $S_t\in \calS_\bad$.
If round $t$ is bad, let \chgred{
$i = \argmin_{j\in \trig{S}_t} N_{j,t-1} \cdot p_j $}. 
If the above $i$ is not unique, we pick an arbitrary one.
Then we increment the counter $N_i$, i.e., $N_{i,t} = N_{i,t-1} + 1$ while not changing other counters
	$N_j$ with $j \ne i$.
If round $t$ is not bad, i.e., $S_t\notin \calS_\bad$, no counter $N_i$ is incremented.
\end{mydef}

Note that the counter $N_i$ is for the purpose of analysis, and its maintenance is not part of the algorithm.
Intuitively, for each round $t$ where a bad super arm $S_t$ is played, 
	we increment exactly one counter $N_i$, where $i$
	is selected among all possibly triggered base arms $\trig{S}_t$ 
	such that the current value of \chgred{ $N_i \cdot p_i$} is the lowest.
In the special case when $p_i=1$ for some $i\in [m]$, we know that
	$i\not \in \trig{S}\setminus S$ for any super arm
	$S$.
Therefore, whenever arm $i$ is selected to increment its counter $N_i$ in a round $t$, $i$ must 
	have been played in round $t$, and thus we have $T_{i,t}\ge N_{i,t}$ 
	for any $i\in [m]$ with $p_i=1$ and all time $t$.
However, this may not holds for $i\in [m]$ with $p_i < 1$, that is, it is possible that
	in a round $t$ a base arm $i$ is not triggered but its counter $N_i$ is incremented.

%
%
%
In every bad round, exactly one counter in $\{N_i\}$
is incremented, so the total number of bad rounds in the first
$n$ rounds is exactly $\sum_{i}N_{i,n}$.
Below we give the definition of refined counters. 

\begin{mydef}[Refined counters]\label{def:refinedcounters}
Each time $N_{i}$ gets updated, one of the bad super arms that could trigger $i$ is played. We further separate $N_i$ into a set of counters as follows:
\[\forall l\in [K_i],\, N_{i,n}^l = \sum_{\chgins{t=1}}^n \I\{S_t = S_{i,\bad}^l, N_{i,t}>N_{i,t-1}\}.\]

That is, each time $N_i$ is updated, we also record
which bad super arm is played. 
\end{mydef}


With these counters in hands, we shall define the two stages
``sufficiently sampled'' and 
``under-sampled'' using the sampling threshold, which 
further split the counter $N_{i,n}^l$ into two 
counters.

\begin{mydef}[Sufficiently sampled and under-sampled]\label{def:sufunder}

Consider time horizon $n$ and current time $t \le n$.
For the refined counter $N_{i,n}^l$'s, we separate them into sufficiently sampled part and under-sampled part, as defined below.
When counter $N_{i,t}^l$ is incremented at time $t$, i.e, $S_t = S_{i,\bad}^l$, we inspect the counter $N_{i,t-1}$. 
If $N_{i,t-1} > \ell_n(\Delta^{i,l},p_i)$, we say that the bad super arm $S_{i,\bad}^l$ is {\em sufficiently sampled} (with respect to base arm $i$);
	otherwise, it is {\em under-sampled} (with respect to base arm $i$). 
Thus counter $N_{i,n}^l$ is separated into the following sufficiently sampled part and under-sampled part:
\begin{align*}
N_{i,n}^{l, suf} =& \sum_{\chgins{t=1}}^n \I\{S_t = S_{i,\bad}^l, N_{i,t}>N_{i,t-1}, N_{i,t-1} > \ell_n(\Delta^{i,l},p_i)\}, \nonumber \\
N_{i,n}^{l, und} =& \sum_{\chgins{t=1}}^n \I\{S_t = S_{i,\bad}^l, N_{i,t}>N_{i,t-1}, N_{i,t-1} \le \ell_n(\Delta^{i,l},p_i)\}.
\end{align*}
\end{mydef}


Following the definition, we have $N_{i,n}^{l,und}\leq \ell_n(\Delta^{i,l},p_i)$, 
and \chgins{$N_{i,n} =\sum_{l\in [K_i]} (N_{i,n}^{l, suf} + N_{i,n}^{l, und})$}. Using this notation, the total reward at time horizon $n$ is at least
\chgins{
\begin{equation}
\label{eqn:reward}
n\cdot \alpha \cdot \opt_{\bp} - \sum_{i\in [m], K_i>0}  \sum_{l\in [K_i]} (N_{i,n}^{l, suf} + N_{i,n}^{l, und}) \cdot \Delta^{i,l}.
\end{equation}}

To get an upper bound on the regret, we want to upper bound 
$N_{i,n}^{l, suf}$ and $ N_{i,n}^{l, und}$ separately. Before doing that, we prove an important connection as follows.

\begin{mylem}[Connection between $N_{i,t-1}$ and $T_{i,t-1}$]
\label{lemma:connection}
Let $n$ be the time horizon.
For every round $t$ with $\chgins{0} < t  \le n$, every base arm $i \in [m]$, every $\Delta >0$, \chgred{ and
	every integer $k > \ell_n(\Delta,p_i)$, we have, 
\begin{equation}
\Pr\left \{ N_{i,t-1} =k, T_{i,t-1} \leq \frac{6\cdot \ln  t}
{f^{-1}(\Delta)^2} \right \}\leq \frac{1}{ t^3}. \label{eq:NandT}
\end{equation}
Moreover, if $p_i$=1, we have 
\[\Pr\left \{ N_{i,t-1} =k, T_{i,t-1} \leq \frac{6\cdot \ln  t}
{f^{-1}(\Delta)^2} \right \}=0.
\]
}
\end{mylem}

\begin{proof}
Fix a base arm $i$. 
The case of $p_i=1$ is trivial since in this case $T_{i,t-1} \ge N_{i,t-1} $ and $n \ge t$. 
Now we only consider the case of $0< p_i < 1$.

In a run of CUCB algorithm (Algorithm~\ref{alg:oracleucb}), let $t^{(j)}$ be the round number 
	at which counter $N_i$ is incremented for the $j$-th time.
Suppose that in round $t^{(j)}$, super arm $S^{(j)}$ is played.
Note that both $t^{(j)}$ and  $S^{(j)}$ are random, depending on the randomness of 
	the outcomes of base arms and the triggering of base arms from super arms in all
	historical rounds.


Let $X^{(j)}$ be the Bernoulli random variable indicating whether
	arm $i$ is triggered by the play of super arm $S^{(j)}$ in round $t^{(j)}$.
\chgins{If in a run of the CUCB algorithm counter $N_i$ is only incremented a finite
	number of times, let $N_{i,\infty}$ denote the final value of the counter $N_i$
	in this run.
In this case, we simply define $X^{(j)}=1$ for all $j> N_{i,\infty}$.
For convenience, when $j> N_{i,\infty}$, we denote the corresponding super arm
	$S^{(j)} = \bot$.
Thus, for any $\ell \ge 1$, $\sum_{j=1}^{\ell} X^{(j)}$ is well defined.
}	
\chgins{For all $0 < t\le n$, 
	since $T_{i,t-1}$ is the number of times $i$ is triggered by the end of round $t-1$, 
	we have
\begin{equation} \label{eq:counter}
	\sum_{j=1}^{N_{i,t-1}} X^{(j)} \le T_{i,t-1}.
\end{equation}
}
	
\chgins{
We now show that for any $j\ge 1$, $\E[X^{(j)} \mid X^{(1)}, \ldots, X^{(j-1)} ]\ge p_i$.
Fixing a super arm $A\in {\cal S}$, if super arm $S^{(j)}$ played in round
	$t^{(j)}$ is $A$, then conditioned on the event $S^{(j)}=A$, in this round whether
	arm $i$ is triggered or not only depends on the randomness of triggering base arms
	after playing $A$, and is independent of randomness in previous rounds. 
In other words, we have 
\begin{equation}
\Pr\left\{X^{(j)}=1 \mid S^{(j)}=A, X^{(1)}, \ldots, X^{(j-1)}\right\}
 = \Pr\left\{X^{(j)}=1 \mid S^{(j)}=A \right\}
 = p_i^A \ge p_i.  \label{eq:pi}
\end{equation}
By the law of total probability, we have
\begin{align}
&\E[X^{(j)} \mid X^{(1)}, \ldots, X^{(j-1)} ] \nonumber \\
& =  \Pr\left\{X^{(j)}=1 \mid X^{(1)}, \ldots, X^{(j-1)} \right\} \nonumber \\
& =  \sum_{A \in {\cal S}} 
 	\Pr\left\{S^{(j)}=A\right\}\cdot \Pr\left\{X^{(j)}=1 \mid S^{(j)}=A, X^{(1)}, \ldots, X^{(j-1)}
 		\right\} 	\nonumber \\
& \ \ \ \ 	+ \Pr\left\{S^{(j)}= \bot\right\}\cdot 
 		\Pr\left\{X^{(j)}=1 \mid S^{(j)}=\bot, X^{(1)}, \ldots, X^{(j-1)}\right\}
 		\nonumber \\
& \ge p_i \sum_{A \in {\cal S}} \Pr\left\{S^{(j)}=A\right\} + 
	\Pr\left\{S^{(j)}= \bot\right\}\cdot 1 
	\label{eq:twoparts}\\
& \ge p_i, \nonumber
\end{align}
where the first part of the Inequality Eq.~\eqref{eq:twoparts} comes from Eq.~\eqref{eq:pi},
	 and the second part comes from our definition that when $S^{(j)}= \bot$, it means
	 that the counter $N_i$ stops before reaching $j$ and $X^{(j)}=1$ in this case.
}

\chgred{
With the result that for any $j\ge 1$, $\E[X^{(j)} \mid X^{(1)}, \ldots, X^{(j-1)} ]\ge p_i$,
	we apply the multiplicative Chernoff bound (Fact~\ref{chernoffLemma}) to
	obtain that for any $\ell \ge 1$, $0<\delta < 1$,
\chgred{
\begin{equation} \label{eq:chernoffpi}
\Pr\left\{\sum_{j=1}^\ell X^{(j)}\leq 
\ell \cdot p_i (1-\delta)
 \right\}\leq e^{-\delta^2 \ell p_i/2}.
\end{equation}
}

We are now ready to carry out the following derivation
	for any $0 < t \le n$, $i \in [m]$, $\Delta >0$, and
		integer $k > \ell_n(\Delta,p_i)$:

\begin{align}
& \Pr\left \{ N_{i,t-1} =k, 
T_{i,t-1}\leq \frac{6\cdot \ln  t}  
{f^{-1}(\Delta)^2} \right \} \nonumber
\\
& \leq  \Pr\left \{ N_{i,t-1} =k,
\sum_{j=1}^{N_{i,t-1}} X^{(j)}
\leq \frac{6\cdot \ln  t}
{f^{-1}(\Delta)^2}  \right \} & \mbox{\{by Eq.~\eqref{eq:counter}\}} \nonumber \\
& \leq  \Pr\left \{\sum_{j=1}^{k} X^{(j)}
\leq \frac{6\cdot \ln  t}
{f^{-1}(\Delta)^2}  \right \}  \nonumber \\
& \leq  \Pr\left \{
\sum_{j=1}^{\lceil \ell_n(\Delta,p_i) \rceil} X^{(j)}
\leq \frac{6\cdot \ln  t}
{f^{-1}(\Delta)^2}  \right \} \nonumber \\
& \leq  \Pr\left \{
\sum_{j=1}^{\lceil \ell_t(\Delta,p_i) \rceil} X^{(j)}
\leq \frac{6\cdot \ln  t}
{f^{-1}(\Delta)^2} \right \}.  \label{eq:lem2_1}
	& \mbox{\{$n\ge t \Rightarrow \ell_n(\Delta,p_i) \ge \ell_t(\Delta,p_i) $\}} 
\end{align}

If $f^{-1}(\Delta)^2\leq \frac12$, let $\delta=\frac12$, we know $\ell_t(\Delta,p_i)=\frac{12\cdot \ln  t}
{f^{-1}(\Delta)^2 \cdot p_i}$, so
\begin{align}
\eqref{eq:lem2_1}& = \Pr\left \{
\sum_{j=1}^{\lceil \ell_t(\Delta,p_i) \rceil} X^{(j)}
\leq \ell_t(\Delta,p_i)\cdot p_i \cdot \frac12 \right \} \nonumber
\\
&\leq \Pr\left \{
\sum_{j=1}^{\lceil \ell_t(\Delta,p_i) \rceil} X^{(j)}
\leq \lceil \ell_t(\Delta,p_i) \rceil \cdot p_i \cdot \frac12 \right \} \nonumber
\\
&\leq e^{-\lceil\ell_t(\Delta,p_i)\rceil p_i /8} & \mbox{\{by Eq.~\eqref{eq:chernoffpi}\}} \nonumber
\\
& \leq e^{-\ell_t(\Delta,p_i) p_i /8}
=e^{-\frac{3\ln  t }{2 f^{-1}(\Delta)^2}}
\leq e^{-3\ln  t  }=\frac{1}{ t^3}. \nonumber
\end{align}

If $f^{-1}(\Delta)^2> \frac12$, 
we know $\ell_t(\Delta,p_i)=\frac{24\cdot \ln  t}
{p_i}$.
Now let $\delta=1-\frac{1}{4f^{-1}(\Delta)^2}\geq \frac12$, which means
$1-\delta=\frac{1}{4f^{-1}(\Delta)^2}$.
 So we have, 

\begin{align*}
\eqref{eq:lem2_1}&= \Pr\left \{
\sum_{j=1}^{\lceil \ell_t(\Delta,p_i) \rceil} X^{(j)}
\leq \frac{24\ln  t}{ 4f^{-1}(\Delta)^2\cdot p_i}\cdot p_i \right \} 
\\
&= \Pr\left \{
\sum_{j=1}^{\lceil \ell_t(\Delta,p_i) \rceil} X^{(j)}
\leq \ell_t(\Delta,p_i)\cdot p_i \cdot (1-\delta) \right \} 
\\&\leq 
e^{-\ell_t(\Delta,p_i) p_i /8} & \mbox{\{by Eq.~\eqref{eq:chernoffpi}\}} \\
&= e^{-3\ln  t  }=\frac{1}{ t^3}.
\end{align*}

Therefore, Inequality~\eqref{eq:NandT} holds.
}
\end{proof}

Recall that a nice run at time $t$ (Definition~\ref{def:nicerun}, denoted as $\calN_t$) 
	means that the difference between the empirical mean and the actual mean
is bounded by the standard difference $\Lambda_{i,t}$ for every arm $i\in [m]$
	\chgins{ 
	($\forall i\in [m], \, |\hat{\mu}_{i,T_{i,t-1}} -\mu_i| \le \Lambda_{i,t}$)}. 
By Lemma \ref{lem:nice}, we know that 
with probability $1-\frac{2m}{t^2}$, $\calN_t$ holds. 
According to
line~\ref{alg:adjustCUCB} of Algorithm~\ref{alg:oracleucb}, we have
\chgins{$\bar{\mu}_{i,t} = \min\{\hat \mu_{i,T_{i,t-1}} + \Lambda_{i,t}, 1\}$}.
Thus, we have 
\begin{align}
\calN_t \Rightarrow \forall i\in [m],\,  \bar{\mu}_{i,t} - \mu_i \,\chgins{\ge}\, 0, 
\label{eqn:differencelowerbound}\\
\calN_t \Rightarrow \forall i\in \trig{S}_t,\, \bar{\mu}_{i,t} -\mu_i \,\chgins{\le}\, 2 \Lambda_t. \label{eqn:differenceupperbound}
\end{align}

Meanwhile, 
by Definition \ref{def:standdif}, we know that for any $i\in [m]$, $l\in [K_i]$ and any time $t$:
\begin{align}\label{eqn:ilget}
\left\{S_t = S_{i,\bad}^l,
\forall s\in \trig{S}_t, T_{s,t-1} >\frac{6\ln  t}{f^{-1}(\Delta^{i,l})^2}
\right\} \Rightarrow \Lambda^{i,l} > \Lambda_t.
\end{align}
With the previous observations, we have the following lemma. Informally, it says that in a nice run in round $t$, it is impossible that the algorithm would select a bad super arm $S_t$ using the oracle, which outputs a correct $\alpha$-approximation answer, while every arm in $\trig{S}_t$ has been tested for enough times.

\begin{mylem}[Impossible case]\label{lem:impossible}
Let $F_t$ be the indicator defined in Definition \ref{def:nonalphaoutput}.
For any $i\in [m]$, $l\in [K_i]$ and any time $t$, the event $\left\{\calN_t, \neg F_t, S_t = S_{i,\bad}^l,
\forall s\in \trig{S}_t, T_{s,t-1} >\frac{6\ln  t}{f^{-1}(\Delta^{i,l})^2}
\right\}$ is empty.
\end{mylem}
\begin{proof}
Indeed, if all the conditions hold, we have:
\begin{align*}
r_{\bp}(S_t) + f(2\Lambda^{i,l}) >& r_{\bp}(S_t) +f(2\Lambda_{t})
&\mbox{\{
  strict monotonicity of $f(\cdot)$ and Eq.(\ref{eqn:ilget})\}}\\
\geq &
r_{\bar{\bp}_t}(S_t)  &\mbox{\{bounded smoothness property and Eq.(\ref{eqn:differenceupperbound})\}}
\\
\geq &\alpha\cdot {\opt_{\bar{\bp}_t}} &
	\mbox{\{$\neg F_t \Rightarrow$  $S_t$ is an $\alpha$
  approximation w.r.t $\bar{\bp}_t$ \}}
\\
\geq &\alpha \cdot r_{\bar{\bp}_t}(S_\bp^*)
&\mbox{\{definition of $\opt_{\bar{\bp}_t}$\}}
\\
\geq & \alpha\cdot r_{\bp}(S_\bp^*)
 = \alpha\cdot \opt_{\bp}. &\mbox{\{ monotonicity of $r_{\bp}(S)$ and Eq.(\ref{eqn:differencelowerbound})\}}
\end{align*}

So we have
\begin{align}
r_{\bp}(S_{i,\bad}^l)+f(2\Lambda^{i,l})> \alpha \cdot \opt_\bp.
 \label{eqn:min_Delta_il}
\end{align}

However, by Definition~\ref{def:standdif},
$f(2\Lambda^{i,l})=f(f^{-1}(\Delta^{i,l}))=\Delta^{i,l}$. Thus, 
Inequality (\ref{eqn:min_Delta_il}) contradicts the definition 
of $\Delta^{i,l}$ in Definition~\ref{def:delta}.
\end{proof}

Now we are ready to prove the bound on sufficiently sampled part. 
Recall that $p^* = \min_{i\in [m]} p_i$.
\begin{mylem}\label{lem:boundonsuffpart}[Bound on sufficiently sampled part]
For any time horizon $n > m$,
\begin{equation}
\label{eqn:suf}
\E\left[\sum_{i\in [m],K_i>0} 
\sum_{l\in [K_i]} N_{i,n}^{l, suf}\right] \leq (1-\beta)n + 
	\chgred{ \frac{(2+\I\{p^* < 1\})m\pi^2}{6}}.
\end{equation}
\end{mylem}
\begin{proof}
%
From Definition~\ref{def:sufunder} on $N_{i,n}^{l,suf}$, we have
\begin{align}
&\E\left[\sum_{i\in [m],K_i>0} 
\sum_{l\in [K_i]} N_{i,n}^{l, suf}\right] \\
&= \E\left[ \sum_{i\in [m],K_i>0} 
\sum_{l\in [K_i]} \sum_{t=1}^n \I\left \{S_t= S_{i,\bad}^l, N_{i,t}>N_{i,t-1}, N_{i,t-1} > \ell_n\left (\Delta^{i,l}, p_i\right )\right \}\right]\nonumber\\
&\le\sum_{i\in [m],K_i>0} 
\sum_{l\in [K_i]}\sum_{t=1}^n \Pr\left \{ S_t = S_{i,\bad}^l, N_{i,t}>N_{i,t-1}, \forall s \in \trig{S}_t, N_{s,t-1} > \ell_n\left (\Delta^{i,l}, \chgred{ p_s}\right ) \right \}, \nonumber 
\end{align}
where the last inequality is due to our way of updating counter $N_i$ by Definition~\ref{def:counters}:
	When $N_i$ is incremented in round $t$ such that $N_{i,t} > N_{i,t-1}$, we know that $N_{i,t-1} \cdot p_i$ has the lowest  
	value among all $N_{s,t-1} \cdot p_s$ for $s\in \trig{S}_t$,
	\chgred{
	and thus $N_{i,t-1} > \ell_n\left (\Delta^{i,l}, p_i\right )$ implies that for $s\in \trig{S}_t$,
	$N_{s,t-1} \ge N_{i,t-1} \cdot p_i /p_s > \ell_n\left (\Delta^{i,l}, p_i\right ) \cdot p_i /p_s = \ell_n\left (\Delta^{i,l}, p_s\right )$.
To prove the lemma, it is sufficient to show that for any $\chgins{0} < t \le n$,
\begin{align}
&\sum_{i\in [m],K_i>0} 
\sum_{l\in [K_i]}\Pr\left \{ S_t = S_{i,\bad}^l, N_{i,t}>N_{i,t-1}, \forall s \in \trig{S}_t, N_{s,t-1} > 
	\ell_n\left (\Delta^{i,l}, \chgred{ p_s}\right ) \right \} \label{eqn:foralls} \\ 
&\leq (1-\beta)+\frac{(2+\I\{p^* < 1\})m}{ t^2}.\label{eqn: N>l}
\end{align}
This is because we may then take the union bound on all $t$'s, and get a bound of 
$$
\sum_{\chgins{t=1}}^n \left( (1-\beta)+\frac{3m}{ t^2} \right) 
	\le (1-\beta)n+\frac{(2+\I\{p^* < 1\})m\pi^2}{6}.
$$
Thus, in order to prove our claim, it suffices to prove Inequality (\ref{eqn: N>l}).

We first split Eq.\eqref{eqn:foralls} into two parts:

\begin{align}
& \sum_{i\in [m],K_i>0} \sum_{l\in [K_i]}
\Pr\left \{ S_t = S_{i,\bad}^l, N_{i,t}>N_{i,t-1}, \forall s \in \trig{S}_t, N_{s,t-1} > 
	\ell_n\left (\Delta^{i,l}, \chgred{ p_s}\right ) \right \} \nonumber\\
=& \sum_{i\in [m],K_i>0} \sum_{l\in [K_i]}
\Pr\left \{ S_t = S_{i,\bad}^l, N_{i,t}>N_{i,t-1}, 
	\forall s \in \trig{S}_t, N_{s,t-1} > \ell_n\left (\Delta^{i,l},p_s\right ) , T_{s,t-1} > \frac{6\cdot \ln  t}
{f^{-1}(\Delta^{i,l})^2}\right \}
\nonumber
\\&+ \sum_{i\in [m],K_i>0} \sum_{l\in [K_i]}
\Pr\left \{ S_t = S_{i,\bad}^l, N_{i,t}>N_{i,t-1}, 
	\forall s \in \trig{S}_t, N_{s,t-1} > \ell_n\left (\Delta^{i,l},p_s\right )
, 	\exists s \in \trig{S}_t, T_{s,t-1} \leq \frac{6\cdot \ln  t}
{f^{-1}(\Delta^{i,l})^2} \right \}. \nonumber \\
=&\Pr\left \{\exists i\in[m], \exists l\in [K_i], S_t = S_{i,\bad}^l, N_{i,t}>N_{i,t-1}, 
	\forall s \in \trig{S}_t, N_{s,t-1} > \ell_n\left (\Delta^{i,l},p_s\right ) , T_{s,t-1} > \frac{6\cdot \ln  t}
{f^{-1}(\Delta^{i,l})^2}\right \}
\nonumber
\\&+ 
\Pr\left \{\exists i\in[m], \exists l\in [K_i], S_t = S_{i,\bad}^l, N_{i,t}>N_{i,t-1}, 
	\forall s \in \trig{S}_t, N_{s,t-1} > \ell_n\left (\Delta^{i,l},p_s\right )
, 	\exists s \in \trig{S}_t, T_{s,t-1} \leq \frac{6\cdot \ln  t}
{f^{-1}(\Delta^{i,l})^2} \right \}, \label{eqn:uniqueil}
\end{align}
where the last equality is due to that the events $\{ S_t = S_{i,\bad}^l, N_{i,t}>N_{i,t-1}\}$ 
	for all $i\in [m]$ and $l\in [K_i]$ are mutually exclusive, since $N_{i,t}>N_{i,t-1}$ determines the unique $i$
	(at most one $N_i$ is incremented in each round by Definition~\ref{def:counters})
	and then $S_t = S_{i,\bad}^l$ determines the unique $l$.

For the first term in Eq.\eqref{eqn:uniqueil}, we apply Lemma \ref{lem:impossible} and have:
\begin{align}
&\forall i\in[m]\, \forall l\in[K_i],\,
\Pr \left\{\calN_t, \neg F_t, S_t=S_{i,\bad}^l, N_{i,t}>N_{i,t-1}, 
\forall s\in \trig{S}_t, \right.
\left. T_{s,t-1} >\frac{6\cdot \ln  t}
{f^{-1}(\Delta^{i,l})^2}
\right\}  = 0 \Rightarrow \nonumber \\
&\Pr\left\{\calN_t, \neg F_t, \exists i\in[m], \exists l\in [K_i],   S_t=S_{i,\bad}^l, N_{i,t}>N_{i,t-1}, 
\forall s\in \trig{S}_t, \right. \left. T_{s,t-1} >\frac{6\cdot \ln  t}
{f^{-1}(\Delta^{i,l})^2}
\right\} = 0 \Rightarrow \nonumber \\
&\Pr \left\{\exists i\in[m], \exists l\in [K_i],   S_t=S_{i,\bad}^l, N_{i,t}>N_{i,t-1}, 
\forall s\in \trig{S}_t, \right. \left. T_{s,t-1} >\frac{6\cdot \ln  t}
{f^{-1}(\Delta^{i,l})^2}
\right\}  \nonumber \\
& \ \ \ \leq \Pr[F_t \lor \neg \calN_t] \leq (1-\beta)+\frac{2m}{ t^2}. \label{eq:FNt}
\end{align}
The inequality in Eq.\eqref{eq:FNt} uses the definition of $F_t$ (Definition~\ref{def:nonalphaoutput}) and Lemma~\ref{lem:nice}.

For the second term in Eq.\eqref{eqn:uniqueil}, we have:
\begin{align}
&\Pr\left \{\exists i\in[m], \exists l\in [K_i], S_t = S_{i,\bad}^l, N_{i,t}>N_{i,t-1}, 
	\forall s \in \trig{S}_t, N_{s,t-1} > \ell_n\left (\Delta^{i,l},p_s\right ), 	
	\exists s \in \trig{S}_t, T_{s,t-1} \leq \frac{6\cdot \ln  t}
{f^{-1}(\Delta^{i,l})^2} \right \}  \nonumber \\
&\le  \Pr\left \{\exists i\in[m], \exists l\in [K_i],
	\exists s \in \trig{S}_t, N_{s,t-1} > \ell_n\left (\Delta^{i,l},p_s\right ), 	
	T_{s,t-1} \leq \frac{6\cdot \ln  t}
{f^{-1}(\Delta^{i,l})^2} \right \}    \nonumber \\
&\le  \sum_{s \in \trig{S}_t} \sum_{k=1}^{t-1}\Pr\left \{\exists i\in[m], \exists l\in [K_i], N_{s,t-1}=k,
	N_{s,t-1} > \ell_n\left (\Delta^{i,l},p_s\right ), 	
	T_{s,t-1} \leq \frac{6\cdot \ln  t}
{f^{-1}(\Delta^{i,l})^2} \right \}.   \label{eqn:fixNtoj} 
\end{align}
Let $\Delta^*(s,k)= \min_{i\in[m], l\in [K_i], \ell_n(\Delta^{i,l},p_s)<k} \Delta^{i,l}$, and $\Delta^*(s,k)=\emptyset$ if the condition
	of $\min$ is not satisfied.
Since $f^{-1}(\Delta)$ decreases when $\Delta$ decreases, we know that when the event
	$\{\exists i\in[m], \exists l\in [K_i], N_{s,t-1}=k,
		N_{s,t-1} > \ell_n\left (\Delta^{i,l},p_s\right ), 	
		T_{s,t-1} \leq \frac{6\cdot \ln  t}
	{f^{-1}(\Delta^{i,l})^2} \}$ is non-empty, it is included in the event
	$\{N_{s,t-1}=k, T_{s,t-1} \leq \frac{6\cdot \ln  t}{f^{-1}(\Delta^*(s,k))^2}  \}$.
Therefore, we have
\begin{align}
\eqref{eqn:fixNtoj} 
& \le \sum_{s \in \trig{S}_t} \sum_{k\in [t-1], \Delta^*(s,k)\ne \emptyset}\Pr\left \{N_{s,t-1}=k, 
	T_{s,t-1} \leq \frac{6\cdot \ln  t}{f^{-1}(\Delta^*(s,k))^2}\right \} \nonumber \\
& \le \sum_{s \in \trig{S}_t} \sum_{k\in [t-1]} \frac{\I\{p_i < 1\}}{ t^3} &\mbox{\{by Lemma~\ref{lemma:connection}\}} \nonumber \\
&\le \frac{\I\{p^* < 1\}m}{t^2}. \label{eqn:splitterm2}
\end{align}
Combining Eq.\eqref{eq:FNt} and Eq.\eqref{eqn:splitterm2}, we obtain Eq.\eqref{eqn: N>l}.
}
\end{proof}

Now we consider 
the bound on under-sampled part, i.e., 
the number of times that
the played bad super arms are under-sampled.
For a particular arm $i$, its counter $N_i$ will increase from \chgins{$0$} to
	$\base_n(\Delta^{i,K_i},p_i)$ before it is sufficiently sampled.
Assume $N_{i,t-1} \in (\base_n(\Delta^{i,j-1},p_i),\base_n(\Delta^{i,j},p_i)]$ when $N_i$ is incremented at time $t$ with an {\em under-sampled} super arm $S_{i,B}^l$. We can conclude that $\Delta^{i,l}\le\Delta^{i,j}$, which will be used as an upper bound for the regret. Otherwise, we must have $\Delta^{i,l} \ge \Delta^{i,j-1}$ and $S_{i,B}^l$ is already {\em sufficiently sampled}.

To simplify the notation, set $\ell_n(\Delta^{i,0},p_i) = 0$.
\chgins{Notice that $N_{i,0}=0$ for all $i$.
For each base arm $i$, the boundary case of $0 = N_{i,t-1} < N_{i,t}$ occurs in only one round
	in a run, and we treat it separately by using $\Delta^{i}_{\max}$ as the regret for this case.
}
%
\chgins{For the rest, we} break the range of the counter $N_{i,t-1}$ \chgins{with $N_{i,t-1}>0$} into discrete segments, i.e., $(\ell_n(\Delta^{i,j-1},p_i), \ell_n(\Delta^{i,j},p_i)]$ for $j\in [K_i]$. Let us assume that the round $t$ is bad and $N_{i,t}$ is incremented. Assume $N_{i,t-1} \in (\ell_n(\Delta^{i,j-1},p_i), \ell_n(\Delta^{i,j},p_i)]$ for some $j$. Notice that we are only interested in the case that $S_t$ is under-sampled.  In particular, if this is indeed the case, $S_t = S_{\bad}^{i,l}$ for some $l \geq j$. (Otherwise, $S_t$ is sufficiently sampled based on the counter $N_{i,t-1}$.) Therefore, we will suffer a regret of $\Delta^{i,l} \leq \Delta^{i,j}$ (See Definition \ref{def:delta}). Consequently, for counter $N_{i,t}$ in range $(\ell_n(\Delta^{i,j-1},p_i), \ell_n(\Delta^{i,j},p_i)]$, we will suffer a total regret for those under-sampled arms at most $(\ell_n(\Delta^{i,j},p_i)-\ell_n(\Delta^{i,j-1},p_i))\cdot \Delta^{i,j}$ in rounds that $N_{i,t}$ is incremented.

\begin{mylem}[Bound on under-sampled part] \label{lem:boundundersampled}
For any time horizon $n>m$, 
we have,
\begin{equation}
\sum_{i\in[m],K_i>0}\sum_{l\in [K_i]} N_{i,n}^{l,und}\cdot \Delta^{i,l}
\leq \sum_{i\in[m],K_i>0}\left (\ell_n(\Delta^i_{\min},p_i) \Delta^i_{\min} + \int_{\Delta^i_{\min}}^{\Delta^i_{\max}} \ell_n(x,p_i) \mathrm{d}x
\chgins{+\Delta^{i}_{\max}}
\right ).\label{eqn:und}
\end{equation}
\end{mylem}

\begin{proof}
It suffices to show that
for any arm $i \in [m]$ with $K_i >0$,
\[\sum_{l\in [K_i]} N_{i,n}^{l,und}\cdot \Delta^{i,l}
\leq 
\ell_n(\Delta^i_{\min},p_i) \Delta^i_{\min} + \int_{\Delta^i_{\min}}^{\Delta^i_{\max}} \ell_n(x,p_i) \mathrm{d}x\chgins{+\Delta^{i}_{\max}}.\]

Now, by definition and discussion on the interval that 
$N_{i,t-1}$ lies in,  we have
\begin{align}
&\sum_{l\in [K_i]} N_{i,n}^{l,und}\cdot \Delta^{i,l} \nonumber \\
& =  \sum_{\chgins{t=1}}^n \sum_{l\in [K_i]}\I\{S_t = S_{i,\bad}^l, N_{i,t}>N_{i,t-1}, N_{i,t-1} \le \ell_n(\Delta^{i,l},p_i)\}\cdot \Delta^{i,l} \nonumber\\
& \chgins{ =  \sum_{\chgins{t=1}}^n \sum_{l\in [K_i]}\I\{S_t = S_{i,\bad}^l, N_{i,t}>N_{i,t-1}, 
	0 < N_{i,t-1} \le \ell_n(\Delta^{i,l},p_i)\}\cdot \Delta^{i,l}} \nonumber \\
& \ \ \ \chgins {+ \sum_{\chgins{t=1}}^n \sum_{l\in [K_i]}\I\{S_t = S_{i,\bad}^l, N_{i,t}>N_{i,t-1}=0 \}\cdot \Delta^{i,l}} \nonumber\\
& \chgins{ \le \sum_{\chgins{t=1}}^n \sum_{l\in [K_i]}\I\{S_t = S_{i,\bad}^l, N_{i,t}>N_{i,t-1}, 
	0 < N_{i,t-1} \le \ell_n(\Delta^{i,l},p_i)\}\cdot \Delta^{i,l} +\Delta^{i}_{\max} } \nonumber \\
& = \sum_{\chgins{t=1}}^n \sum_{l\in [K_i]} \sum_{j=1}^l \I\{S_t = S_{i,\bad}^l, N_{i,t}>N_{i,t-1}, N_{i,t-1} \in (\ell_n(\Delta^{i,j-1},p_i), \ell_n(\Delta^{i,j},p_i) ] \} \cdot \Delta^{i,l} \chgins{+\Delta^{i}_{\max}}\nonumber\\
& \le \sum_{\chgins{t=1}}^n \sum_{l\in [K_i]} \sum_{j=1}^{l} \I\{S_t = S_{i,\bad}^l, N_{i,t}>N_{i,t-1}, N_{i,t-1} \in (\ell_n(\Delta^{i,j-1},p_i), \ell_n(\Delta^{i,j},p_i) ] \} \cdot \Delta^{i,\boldsymbol{j}}\chgins{+\Delta^{i}_{\max}} \label{eqn:undersplit} \\
&\le \sum_{\chgins{t=1}}^n \sum_{l\in [K_i]} \sum_{j\in [K_i]} \I\{S_t = S_{i,\bad}^l, N_{i,t}>N_{i,t-1}, N_{i,t-1} \in (\ell_n(\Delta^{i,j-1},p_i), \ell_n(\Delta^{i,j},p_i) ] \} \cdot \Delta^{i,{j}}\chgins{+\Delta^{i}_{\max}} \nonumber\\
&= \sum_{\chgins{t=1}}^n \sum_{j\in [K_i]} \I\{S_t \in \calS_{i,\bad}, N_{i,t}>N_{i,t-1}, N_{i,t-1} \in (\ell_n(\Delta^{i,j-1},p_i), \ell_n(\Delta^{i,j},p_i) ] \} \cdot \Delta^{i,{j}}\chgins{+\Delta^{i}_{\max}}. \label{eqn:undermerge}
\end{align}
The Inequality~\eqref{eqn:undersplit} holds since $\Delta^{i,j}\geq \Delta^{i,l}$
for $j\leq l$. 
Equality~\eqref{eqn:undermerge} is by first switching summations and
	then merging all $S_{i,\bad}^l$ into $\calS_{i,\bad}$.
We may now switch the summations again, and get
\begin{align}
(\ref{eqn:undermerge})=&  \sum_{j\in [K_i]} \sum_{\chgins{t=1}}^n\I\{S_t \in \calS_{i,\bad}, N_{i,t}>N_{i,t-1},  N_{i,t-1} \in (\ell_n(\Delta^{i,j-1},p_i), \ell_n(\Delta^{i,j},p_i) ] \} \cdot \Delta^{i,j}\chgins{+\Delta^{i}_{\max}} \nonumber \\
\chgred{\leq}& \chgred{\sum_{j\in [K_i]} (\lfloor \ell_n(\Delta^{i,j},p_i)\rfloor - \lfloor \ell_n(\Delta^{i,j-1},p_i)\rfloor)\cdot \Delta^{i,j}\chgins{+\Delta^{i}_{\max}}}
 \label{eqn:underrelax} \\
\chgred{\leq} & \chgred{\sum_{j\in [K_i]} (\ell_n(\Delta^{i,j},p_i) - \ell_n(\Delta^{i,j-1},p_i))\cdot \Delta^{i,j}\chgins{+\Delta^{i}_{\max}}.}
 \label{eqn:underrelax2}
\end{align}
Inequality~\eqref{eqn:underrelax} uses a relaxation on the indicators. 
\chgred{ In Inequality~\eqref{eqn:underrelax2}, for every $j\ge 2$, we relax the part of 
	$(\ell_n(\Delta^{i,j-1},p_i) - \lfloor \ell_n(\Delta^{i,j-1},p_i)\rfloor)\cdot \Delta^{i,j}$ to
	$(\ell_n(\Delta^{i,j-1},p_i) - \lfloor \ell_n(\Delta^{i,j-1},p_i)\rfloor)\cdot \Delta^{i,j-1}$.}
Now we simply expand the summation, and some terms will be cancelled. Then, we upper bound the new summation using an integral:
\begin{align}
\eqref{eqn:underrelax2}=& \ell_n(\Delta^{i,K_i},p_i) \Delta^{i,K_i}+\sum_{j\in [K_i-1]} \ell_n(\Delta^{i,j},p_i) \cdot( \Delta^{i,j} -\Delta^{i,j+1})\chgins{+\Delta^{i}_{\max}}
\nonumber \\
\leq& \ell_n(\Delta^{i,K_i},p_i) \Delta^{i,K_i} + \int_{\Delta^{i,K_i}}^{\Delta^{i,1}} \ell_n(x,p_i) \mathrm{d}x\chgins{+\Delta^{i}_{\max}} \label{eq:integralinq} \\
= & \ell_n(\Delta^i_{\min},p_i) \Delta^i_{\min} + \int_{\Delta^i_{\min}}^{\Delta^i_{\max}} \ell_n(x,p_i) \mathrm{d}x\chgins{+\Delta^{i}_{\max}}.
\label{eq:tightint}
\end{align}
Inequality~\eqref{eq:integralinq} comes from the fact that $\ell_n(x,p_i)$ is decreasing in $x$.
\end{proof}

Finally we are ready to prove our main theorem. We just need to combine the upper bounds from the sufficiently sampled part and the under-sampled part together. 

\begin{proof}[Proof of Theorem~\ref{thm:cucb individual}]
Using the counters defined in Definition \ref{def:refinedcounters}, 
we may get the expectation of the regret by 
computing the expectation of the value of the counters
after the $n$-th round. More specifically, according to Definition 
\ref{def:regretdef}, the expected regret is the difference between 
$n \cdot \alpha  \cdot
\beta\cdot  \opt_{\bp}$ and the expected
reward, which is at least $\alpha \cdot n\cdot \opt_{\bp}$ minus
the expected loses from playing bad super arms.

Therefore, combining with Eq.(\ref{eqn:suf}) and Eq.(\ref{eqn:und}), the overall regret of our algorithm is
\begin{align}
& Reg^A_{\bp,\alpha,\beta}(n) \nonumber \\
& \leq \E \left[n \cdot \alpha  \cdot
\beta\cdot  \opt_{\bp}  -
\left(\alpha \cdot n\cdot \opt_{\bp} - \sum_{i\in[m], K_i>0} 
\left(
\sum_{l\in [K_i]} (N_{i,n}^{l, suf} + N_{i,n}^{l, und}) \cdot \Delta^{i,l}\right)\right)\right]\label{eqn:combinetwoparts}\\
& \leq \Delta_{\max} \cdot \E \left[ \sum_{i\in[m], K_i>0} \sum_{ l\in [K_i]} N_{i,n}^{l, suf}\right]  \nonumber
\\
& \ \ \ 
+ \sum_{i\in[m], K_i>0} \left( \ell_n(\Delta^i_{\min},p_i) \Delta^i_{\min} +  \int_{\Delta^i_{\min}}^{\Delta^i_{\max}} \ell_n(x,p_i) \mathrm{d}x\chgins{+\Delta^{i}_{\max}}\right) 
	- (1-\beta)\cdot n\cdot \alpha \cdot \opt_{\bp}\nonumber\\
& \leq
 \sum_{i\in[m], K_i>0}\left( 
  \ell_n(\Delta^i_{\min},p_i) \Delta^i_{\min} +  \int_{\Delta^i_{\min}}^{\Delta^i_{\max}} \ell_n(x,p_i) \mathrm{d}x\right) 
 + \chgred{\left(\frac{(2+\I\{p^* < 1\})\pi^2}{6}+1\right)\cdot m\cdot \Delta_{\max} }
 \nonumber\\
& \ \ \ 
 +(1-\beta)n \cdot \Delta_{\max}
- (1-\beta)\cdot n\cdot \alpha \cdot \opt_{\bp}\label{eq:neednonnegative}\\
& \leq  \sum_{i\in[m], K_i>0}\left(  \ell_n(\Delta^i_{\min},p_i) \Delta^i_{\min} +  \int_{\Delta^i_{\min}}^{\Delta^i_{\max}} \ell_n(x,p_i) \mathrm{d}x
 \right)
+ \chgred{ \left(\frac{(2+\I\{p^* < 1\})\pi^2}{6}+1\right)\cdot m\cdot \Delta_{\max}}.  \label{eq:regretend} 
\end{align}
The last step of derivation from Eq.\eqref{eq:neednonnegative} to Eq.\eqref{eq:regretend} uses the fact that all rewards are
	nonnegative and thus $\Delta_{\max} \le \alpha \cdot \opt_{\bp}$ by Definition~\ref{def:delta}.
\end{proof}

\subsubsection{Proof of Theorem~\ref{thm:worstbound}}

The proof of Theorem~\ref{thm:worstbound} relies on the tight regret bound for the leading $\ln n$ term given by Theorem~\ref{thm:cucb individual}.

\begin{proof}[Proof of Theorem~\ref{thm:worstbound}]
We first prove the case of $p^*=1$.
Following the proof of Theorem~\ref{thm:cucb individual}, we only need to consider the base arms that are played when they are under-sampled. 
Following the intuition, we need to quantify when $\Delta$ is too small. In particular, we measure the threshold for $\Delta^i_{\min}$ based on $N_{i,n}$, i.e., the counter of arm $i$ at time horizon $n$.
Let $\{n_j \mid j \in [m]\}$ be a set of possible counter values at time horizon $n$. Our analysis will then be conditioned on event  ${\cal E} = \{\forall j\in [m], N_{j,n} = n_j\}$. 

For an arm $i\in [m]$ with $K_i > 0$, we have
\begin{align*}
&\sum_{l\in [K_i]} N_{i,n}^{l,und}\cdot \Delta^{i,l}
\mid {\cal E} 
\\=& \sum_{\chgins{t=1}}^n \sum_{l\in [K_i]}\I\{S_t = S_{i,\bad}^l, N_{i,t}>N_{i,t-1}, N_{i,t-1} \le \ell_n(\Delta^{i,l},1)\, \mid {\cal E}  \}\cdot \Delta^{i,l} \nonumber
\end{align*}
With $f(x) = \gamma x^\omega$, we have $f^{-1}(x) = \left(\frac{x}{\gamma}\right)^{1/\omega}$.
Define $\Delta^*(n_i) = \left(\frac{6\gamma^{2/\omega} \ln  n}{n_i}\right)^{\omega/2}$, i.e., $\base_n(\Delta^*(n_i), 1) = n_i$.
Now we consider two cases.

Case (1): $\Delta_{\min}^i  > \Delta^*(n_i)$. Following the same derivation as in the proof of Lemma~\ref{lem:boundundersampled}
	(notice that the same derivation still works when conditioned on event $\cal E$),
	we have
\begin{align}
\sum_{l\in [K_i]} N_{i,n}^{l,und}\cdot \Delta^{i,l} \mid {\cal E} & \leq 
	 \ell_n(\Delta^i_{\min},1) \Delta^i_{\min} + \int_{\Delta^i_{\min}}^{\Delta^i_{\max}} \ell_n(x,1) \mathrm{d}x \chgins{+\Delta^{i}_{\max}}  \label{eq:ponecase1b}\\
	 & = \frac{6\gamma^{\frac{2}{\omega}} \ln  n}{(\Delta^i_{\min})^{\frac{2}{\omega}-1}} 
	 	+ \frac{\omega}{2-\omega} 6 \gamma^{\frac{2}{\omega}} 
	 	\ln  n\left((\Delta^i_{\min})^{1-\frac{2}{\omega}} -
	 	 (\Delta^i_{\max})^{1-\frac{2}{\omega}} \right) \chgins{+\Delta^{i}_{\max}}
	 	 \nonumber \\
	 & \le \frac{2}{2-\omega}\cdot
	 \frac{6 \cdot \gamma^{\frac{2}{\omega}} \ln  n}{ (\Delta_{\min}^i)^{\frac{2}{\omega}-1}}
	 \leq
	 \frac{2\gamma}{2-\omega}\cdot (6\ln  n)^{\omega/2}n_i^{1-\omega/2} \chgins{+\Delta^{i}_{\max}}. \label{eq:ponecase1e}
\end{align}	
The last inequality above is by replacing $\Delta_{\min}^i$ with $\Delta^*(n_i)$.

Case (2): $\Delta_{\min}^i \leq \Delta^*(n_i)$.
Let $l^* = \min\{ l\in [K_i] \mid \Delta^{i,l} \le \Delta^*(n_i) \}$.
Notice that $\Delta^{i,l^*} \leq \left(\frac{6\gamma^{2/\omega} \ln  n}{n_i}\right)^{\omega/2}$.
We follow the 
same derivation as in the proof of Lemma~\ref{lem:boundundersampled},
and then 
we critically use the fact that the counter $N_i$ cannot go beyond $n_i$ (in the first term in 
	Inequality~\eqref{eq:criticalni}):
\begin{align}
&\sum_{l\in [K_i]} N_{i,n}^{l,und}\cdot \Delta^{i,l}
\mid {\cal E}\nonumber\\
\le &  \sum_{j\in [K_i]} \sum_{\chgins{t=1}}^n\I\{S_t \in \calS_{i,\bad}, N_{i,t}>N_{i,t-1},  N_{i,t-1} \in (\ell_n(\Delta^{i,j-1},1), \ell_n(\Delta^{i,j},1) ] \mid {\cal E} \} \cdot \Delta^{i,j}  \chgins{+\Delta^{i}_{\max}} \label{eq:ponecase2b} \\
\le & \sum_{j \ge l*} \sum_{\chgins{t=1}}^n\I\{S_t \in \calS_{i,\bad}, N_{i,t}>N_{i,t-1},  N_{i,t-1} \in (\ell_n(\Delta^{i,j-1},1), \ell_n(\Delta^{i,j},1) ] \mid {\cal E}\} \cdot \Delta^*(n_i) 
\chgins{+\Delta^{i}_{\max}} \nonumber \\
	& + \sum_{j \in [l*-1]} \sum_{\chgins{t=1}}^n\I\{S_t \in \calS_{i,\bad}, N_{i,t}>N_{i,t-1},  N_{i,t-1} \in (\ell_n(\Delta^{i,j-1},1), \ell_n(\Delta^{i,j},1) ] \mid {\cal E} \} \cdot \Delta^{i,j}  \chgins{+\Delta^{i}_{\max}}\nonumber \\
\le & (n_i - \ell_n(\Delta^{i,l^*-1},1))\cdot \Delta^*(n_i)  
	+ \sum_{j\in [l^*-1]} (\ell_n(\Delta^{i,j}, 1) - \ell_n(\Delta^{i,j-1}, 1))\cdot \Delta^{i,j}
	\chgins{+\Delta^{i}_{\max}} \label{eq:criticalni}\\
\leq& n_i \cdot \Delta^*(n_i) + \int_{\Delta^*(n_i)}^{\Delta^{i,1}} \ell_n(x, 1) \mathrm{d}x  
	\chgins{+\Delta^{i}_{\max}}
\leq 
\frac{2\gamma}{2-\omega}\cdot (6\ln  n)^{\omega/2}n_i^{1-\omega/2}
	\chgins{+\Delta^{i}_{\max}}.\label{eqn:und2}
\end{align}
Therefore, Eq.(\ref{eqn:und2}) holds in both cases. We then have
\begin{align}
\sum_{i\in [m], K_i > 0}\sum_{l\in [K_i]} N_{i,n}^{l,und}\cdot \Delta^{i,l}
\mid {\cal E}
&\leq
\frac{2\gamma}{2-\omega}\cdot (6\ln  n)^{\omega/2} \cdot \sum_{i\in [m], K_i > 0} n_i^{1-\omega/2} \chgins{+\Delta^{i}_{\max}}
	\nonumber\\
&\leq
\frac{2\gamma}{2-\omega}\cdot (6m\ln  n)^{\omega/2} \cdot n^{1-\omega/2}\chgins{+\Delta^{i}_{\max}}.\label{eq:underInd}
\end{align}

The last inequality comes from Jensen's inequality and $\sum_i n_i \leq n$.
Since the final inequality does not depend on $n_i$, we can drop the condition $\cal E$ above.
With the bound on the under-sampled part given in Inequality~\eqref{eq:underInd}, we combine it
	with the result on sufficiently sampled part given in Lemma~\ref{lem:boundonsuffpart},
	then we can following the similar derivation as
		shown from Eq.\eqref{eqn:combinetwoparts} to Eq.\eqref{eq:regretend}
		to derive the distribution-independent regret bound given in
	Theorem~\ref{thm:worstbound} for the case of $p^*=1$.
	
We now prove the case of $p^* < 1$.
The proof is essentially the same, but with a different definition of $\ell_n(\Delta,p)$.
\chgred{ For convenience, we relax $\ell_n(\Delta,p) = \max\left (\frac{12\cdot \ln  n}{(f^{-1}(\Delta))^2\cdot p}, 
\frac{24\cdot \ln  n}{p} \right )$ to $\frac{12\cdot \ln  n}{(f^{-1}(\Delta))^2\cdot p} + \frac{24\cdot \ln  n}{p}$.}
In this case, we define $\Delta^*_i(n_i) = \left(\frac{12\gamma^{2/\omega} \ln  n}{p_i n_i}\right)^{\omega/2}$.

For Case (1): $\Delta_{\min}^i  > \Delta^*_i(n_i)$, following the same derivation as
	Eq.\eqref{eq:ponecase1b}-\eqref{eq:ponecase1e} except that we use $\ell_n(\cdot,p_i)$
	instead of $\ell_n(\cdot,1)$ (Definition~\ref{def:samplthreshold}), we have
\[
\sum_{l\in [K_i]} N_{i,n}^{l,und}\cdot \Delta^{i,l} \mid {\cal E}\le 
	\frac{2\gamma}{2-\omega}\cdot \left(\frac{12\ln  n}{p_i}\right)^{\omega/2}n_i^{1-\omega/2} + 
	\chgred{\frac{24\ln  n}{p_i} } \cdot \Delta^i_{\max} \chgins{+\Delta^{i}_{\max}}.
\]
For Case (2): $\Delta_{\min}^i \leq \Delta^*_i(n_i)$, again following the same derivation
	Eq.\eqref{eq:ponecase2b}-\eqref{eqn:und2} except that we use $\ell_n(\cdot,p_i)$
		instead of $\ell_n(\cdot,1)$,, we have
\[
\sum_{l\in [K_i]} N_{i,n}^{l,und}\cdot \Delta^{i,l} \mid {\cal E} \le 
	\frac{2\gamma}{2-\omega}\cdot \left(\frac{12\ln  n}{p_i}\right)^{\omega/2}n_i^{1-\omega/2} + 
	\chgred{\frac{24\ln  n}{p_i} } \cdot \Delta^i_{\max} \chgins{+\Delta^{i}_{\max}}.
\]
Together, we have
\begin{align*}
&\sum_{i\in [m], K_i > 0}\sum_{l\in [K_i]} N_{i,n}^{l,und}\cdot \Delta^{i,l} \mid {\cal E} \\
& \le 
	\frac{2\gamma}{2-\omega}\cdot \left(\frac{12\ln  n}{p^*}\right)^{\omega/2} \sum_{i\in [m], K_i > 0} n_i^{1-\omega/2} 
	+ \sum_{i\in [m], K_i > 0} \chgred{\frac{24\ln  n}{p_i}} \cdot \Delta^i_{\max} \\
& \le 
	\frac{2\gamma}{2-\omega}\cdot \left(\frac{12 m \ln  n}{p^*}\right)^{\omega/2} n^{1-\omega/2} 
	+  \sum_{i\in [m]} \chgred{\frac{24\ln  n}{p_i} } \cdot \Delta_{\max} \chgins{+\Delta^{i}_{\max}}.
\end{align*}
Finally, combining Lemma~\ref{lem:boundonsuffpart} and the derivation for the regret bound as
	shown from Eq.\eqref{eqn:combinetwoparts} to Eq.\eqref{eq:regretend}, we obtain the regret bound for the case of $p^* < 1$.
\end{proof}

\subsection{Discussions}

We may further improve the bound in Theorem \ref{thm:cucb individual} as follows, when all the triggering probabilities
are $1$. 

\paragraph{Improving the coefficient of the leading term when $\forall i, p_i=1 $.}
In general, we can
set $\bar{\mu}_i = \hat{\mu}_i +\sqrt{y/(2T_{i})}$ for some $y$ in 
line~6 in the CUCB algorithm. 
The corresponding regret bound obtained is
\[
\sum_{i\in [m], K_i>0} \left( \frac{2\cdot y}{(f^{-1}(\Delta_{\min}^i))^2} \cdot \Delta^{i}_{\min} + \int_{\Delta^{i}_{\min}}^{\Delta^{i}_{\max}} \frac{2\cdot y}{(f^{-1}(x))^2} \mathrm{d}x \right)
+ \left(1 + \sum_{\chgins{t=1}}^n\frac{2t}{e^{-y}} \right)\cdot m \cdot \Delta_{\max}.
\]
What we need is to make sure the term $\sum_{\chgins{t=1}}^n\frac{2t}{e^{-y}}$ in the above
	regret bound converges.
We can thus set $y$ appropriately to guarantee convergence while improving the constant
	in the leading term.
One way is setting $y=(1+c)\ln t$ with a constant $c>1$,
	or equivalently setting $\bar{\mu}_i = \hat{\mu}_i +\sqrt{(1+c)\ln t/(2T_{i})}$, so that $\sum_{\chgins{t=1}}^n\frac{2t}{e^{-y}} = 2\sum_{\chgins{t=1}}^n t^{-c}
	\le 2\zeta(c) $, where $\zeta(c)=\sum_{t=1}^\infty \frac{1}{t^{c}}$ is the Riemann's zeta function, and has a finite value when $c>1$.
Then the regret bound is
\[
\sum_{i\in [m],  K_i>0 } \left( \frac{2\cdot(1+c)\cdot\ln n}{(f^{-1}(\Delta_{\min}^i))^2} \cdot \Delta^{i}_{\min} + \int_{\Delta^{i}_{\min}}^{\Delta^{i}_{\max}} \frac{2\cdot(1+c)\cdot\ln n}{(f^{-1}(x))^2} \mathrm{d}x \right)
+ (2\cdot \zeta(c)+1)\cdot m \cdot \Delta_{\max}.
\]

We can also further improve the constant factor from $2(1+c)$ to $4$ by
	setting $\bar{\mu}_i = \hat{\mu}_i +\sqrt{\frac{2\ln t + \ln\ln t}{2T_{i}}}$ at the cost of a second order $\ln\ln n$ term as in~\cite{Garivier2011}, with regret at most
\[
\sum_{i\in [m],  K_i>0} \left( \frac{2 \cdot(2\ln n +\ln\ln n)}{(f^{-1}(\Delta_{\min}^i))^2} \cdot \Delta^{i}_{\min} + \int_{\Delta^{i}_{\min}}^{\Delta^{i}_{\max}} \frac{2\cdot(2\ln n +\ln\ln n)}{(f^{-1}(x))^2} \mathrm{d}x \right)
+ (1+2\ln\ln n)\cdot m \cdot \Delta_{\max}.
\]

This is because $\sum_{\chgins{t=1}}^n\frac{1}{t\ln t}\leq \int_{m}^{n} \frac{1}{t\ln t} \mathrm{d}t\leq \ln\ln n$ when $m >e$.

\paragraph{Comparing to classical MAB.}
As we discussed earlier, the classical MAB is a special instance of our	
	CMAB framework in which each super arm is a simple arm, $p_i=1$ for all $i\in [m]$,
	function $f(\cdot)$ is the identity function, and
	$\alpha=\beta=1$.
Notice that $\Delta_{\max}^i = \Delta_{\min}^i$.
Thus, by Theorem~\ref{thm:cucb individual}, the regret bound of the classical
	MAB is
\begin{equation} \label{eq:mabnew}
\sum_{i\in [m],\Delta^i>0} \frac{6\ln n
 }{\Delta^i}
+ \left( \frac{\pi^2}{3}+1
\right) \cdot m \cdot
\Delta_{\max},
\end{equation}
where $\Delta^i=\max_{j\in [m]} \mu_j-\mu_i$.
Comparing with the regret bound in Theorem~1 of~\cite{AuerCF02}, we see that we even have
	a better coefficient $\sum_{i\in [m],\Delta^i>0} 6/\Delta^i$ in the leading
	$\ln n$ term than the one $\sum_{i\in [m],\Delta^i>0} 8/\Delta^i$ in the original analysis of UCB1.\footnote{We remark 
	that the constant of UCB1 has been tightened to the optimum~\citep{Garivier2011}.}
The improvement is due to a tighter analysis, and is the reason that we obtained improved regret over~\cite{Yi2012}.
\chgins{Thus, the regret upper bound of our CUCB algorithm when applying to the classical MAB problem is at the same
	level (up to a constant factor) as UCB1,
	which is designed specifically for the MAB problem.}

\vspace{\sectionspace}
\section{Applications}
\label{sec:app}

In this section, we describe two applications with non-linear reward functions as well as the class of linear reward applications that fit our CMAB framework. Notice that, the probabilistic maximum coverage bandit and social influence maximization bandit are also instances of the online submodular maximization problem, which can be addressed in the adversarial setting by~\cite{Streeter2008}, but we are not aware of
	their counterpart in the stochastic setting.

\vspace{\subsectionspace}
\subsection{Probabilistic maximum coverage bandit}

The online advertisement placement application discussed in the
	introduction can be modeled by the
	bandit version of the
	probabilistic maximum coverage (PMC) problem.
PMC has as input a weighted bipartite graph $G=(L,R,E)$ where each edge
	$(u,v)$ has a probability $p(u,v)$, and it needs to find a set
	$S\subseteq L$ of size $k$ that maximizes the expected number of
	activated nodes in $R$, where a node $v\in R$ can be activated by a node
	$u\in S$ with an independent probability of $p(u,v)$.
In the advertisement placement scenario, $L$ is the set of web pages, $R$
	is the set of users, and $p(u,v)$ is the probability that user $v$ clicks the
	advertisement on page $u$.
PMC problem is NP-hard, since when all edge probabilities
	are $1$, it becomes the NP-hard Maximum Coverage problem.

Using submodular set function maximization technique~\citep{NWF78},
	it can be easily shown
	that there exists a deterministic $(1-1/e)$ approximation algorithm
	for the PMC problem, which means that we have a $(1-1/e,1)$-approximation
	oracle for PMC.

The PMC bandit problem is that edge probabilities are unknown, and
 	one repeatedly selects $k$ targets in $L$ in multiple rounds,
	observes all edge activations and adjusts target selection accordingly
	in order to maximize the total number of activated nodes over
	all rounds.

We can formulate this problem as an instance in the CMAB framework.
Each edge $(u,v)\in E$ represents an arm, and each play of the arm
	is a $0$-$1$ Bernoulli random variable with parameter $p_{u,v}$.
A super arm is the set of edges $E_S$ incident to a set
	 $S\subseteq L$ of size $k$.
The reward of $E_S$ is the number of activated nodes in $R$, which is the number of
	nodes in $R$ that are incident to at least one edge in $E_S$ with outcome $1$.
Since all arms are independent Bernoulli random variables, we know that the expected 
	reward only depends on the probabilities on all edges.
In particular we have that the expected reward $r_{\bp}(E_S)=\sum_{v\in R} (1-\prod_{u\in L, (u,v)\in E_S}(1-p(u,v)))$.
Note that this expected reward function is not linear in $\bp=\{p(u,v)\}_{(u,v)\in E}$.
For all arm $i\in E$, we have $p_i=1$, that is, we do not have probabilistically triggered arms.
The monotonicity property is straightforward.
The bounded smoothness function is $f(x) = |E|\cdot x$,
	i.e., increasing all probabilities of all arms in a super arm
	by $x$ can increase the expected
	number of activated nodes in $V$ by at most $|E|\cdot x$.
Since $f(\cdot)$ is a linear function, the integral in Eq.(\ref{eqn:detailed regret bound})
	has a closed form.
In particular, by Theorem~\ref{thm:cucb individual}, we know that
	the distribution-dependent $(1-1/e,1)$-approximation regret bound of our CUCB algorithm
	on PMC bandit
    is \vspace{-1mm}
\[
\sum_{i\in E, K_i>0} \frac{12\cdot |E|^2\cdot \ln n}{\Delta^i_{\min}}
+\left(\frac{\pi^2}{3}+1\right)\cdot |E| \cdot \Delta_{\max}.
\]

Notice that all edges incident to a node $u\in L$ are always played together. In other words, these edges can share one counter. We call these arms (edges) as {\em clustered arms}. It is possible to exploit this property to improve the coefficient of the $\ln n$ term, so that the summation is not among all edges but only nodes in $L$. (See Section 4.1 and the supplementary material of~\cite{CWY13} for the regret bound and analysis for the case of clustered arms).

From Theorem~\ref{thm:worstbound}, we also have the distribution-independent regret bound of 
\[
\sqrt{24 |E|^{3} n \ln n} + \left(\frac{\pi^2}{3}+1\right)\cdot |E| \cdot \Delta_{\max}.
\]
Note that for the PMC bandit, $\Delta_{\max}$ is at most the number of
	vertices covered in $R$, and thus $\Delta_{\max} \le |R|$.

\vspace{\subsectionspace}
\subsection{Combinatorial bandits with linear rewards}
\label{sec:linear}

\cite{Yi2012} studied the {\em Learning with Linear Reward} policy (LLR).
Their formulation is close to ours except that their reward function must be linear.
In their setting, there are $m$ underlying arms.
There are a finite number of super arms, each of which
	consists of a set of underlying arms $S$ together
with a set of coefficients $\{w_{i,S} \mid i\in S\}$.
The reward of playing super arm $S$ is $\sum_{i\in S}w_{i,S} \cdot X_i$, where $X_i$ is the
	random outcome of arm $i$.
The formulation can model a lot of bandit problems appeared in the literature, e.g., multiple plays, shortest path,
	minimum spanning tree and maximum weighted matching.

Our framework contains such linear reward problems as special cases.\footnote{To include
the linear reward case, we allow two super arms with the same set of underlying arms
to have different sets of coefficients. This is fine as long as the oracle
could output super arms with appropriate parameters.}
In particular, let $L=\max_S |S|$ and $a_{\max}=\max_{i,S} w_{i,S}$, and
	we have the bounded smoothness function $f(x) = a_{\max}\cdot L \cdot x$.
In this setting we have $p_i=1$ for all $i\in [m]$.
By applying Theorem~\ref{thm:cucb individual}, the regret bound is
\[
\left(
\sum_{ i\in[m],  K_i>0}\frac{12\cdot a_{\max}^2\cdot L^2 \cdot \ln n}{\Delta_{\min}^i}\right)
+ \left(\frac{\pi^2}{3}+1\right)\cdot m \cdot \Delta_{\max}.
\]
Our result significantly improves the coefficient of the leading $\ln n$ term
	comparing to Theorem 2 of~\citep{Yi2012} in two aspects:
	(a) we remove a factor of $L+1$;
	and
	(b) the coefficient $\sum_{ i\in[m], \Delta_{\min}^i >0}1/\Delta_{\min}^i$ is likely to
		be much smaller than $m\cdot\Delta_{\max}/(\Delta_{\min})^2$ in~\citep{Yi2012}.
This demonstrates that while our framework covers a much larger class of problems, we are
	still able to provide much tighter analysis than the one for linear reward bandits.
Moreover, applying Theorem~\ref{thm:worstbound} we can obtain distribution-independent bound for combinatorial bandits with linear
	rewards, which is not provided in~\citep{Yi2012}:
\[
 a_{\max} L \sqrt{24m n \ln n} + \left(\frac{\pi^2}{3}+1\right)\cdot m \cdot \Delta_{\max}.
\]
Note that, for the class of linear bandits, the reward is at most $a_{\max}\cdot L$, and thus $\Delta_{\max}\le a_{\max}\cdot L$.

\chgins{
We remark that, in a latest paper, \citet{KWAS15} show that the above regret bounds
	can be improved to $O(L \log n \sum_i 1 / \Delta_{\min}^i)$ for distribution-dependent
	regret and $O(\sqrt{L m n \log n})$ for distribution-independent regret, respectively, 
	which are tight (up to a factor of $\sqrt{\log n}$ for the distribution-independent bound).
The improvement is achieved by a weaker and non-uniform sufficient sampling condition ---
	in our analysis, we require all relevant base arms of a super arm $S_t$ played in round $t$
	to be sufficiently sampled to ensure that $S_t$ cannot be a bad super arm (Lemma~\ref{lem:impossible}), but in \citep{KWAS15}, they relax this and show that
	it is enough to have sufficiently many base arms to be sampled sufficiently many times, while
	the rest arms only need to satisfy some weaker sufficient sampling condition.
The intuition is that due to linear reward summation, as long as many base arms are
	sufficiently sampled and the rest have a weaker sufficiently sampled condition, the sum of
	the errors would be still small enough to guarantee that a good super arm is selected
	by the oracle.
However, it is unclear if this technique can be applied to non-linear reward functions satisfying
	our bounded smoothness assumption, since the estimate error of each base arm may not linearly
	affect the estimate error in the expected reward.
}

\subsection{Application to social influence maximization} \label{sec:infmax}
In social influence maximization with the independent cascade model \citep{kempe03},
	we are given a directed graph
	$G=(V,E)$, where every edge $(u,v)$
	is associated with an unknown {\em influence
	probability} $p_{u,v}$.
Initially, a seed set $S\subseteq V$ are selected and activated.
In each iteration of the diffusion process, each node $u$ activated in the
	previous iteration has one chance of activating
	its inactive outgoing neighbor $v$ independently
	with probability $p_{u,v}$.
The reward of $S$ after the diffusion process is
	the total number of activated nodes in the end.
Influence maximization is to find a seed set $S$ of at most $k$ nodes
	that maximize the expected reward, also referred to as the {\em influence spread } of seed set $S$.
\cite{kempe03} show that the problem is NP-hard and
	provide
	an algorithm
	with approximation ratio $1-1/e-\varepsilon$ with
	success probability ($1-1/|E|$)
	for any fixed $\varepsilon >0$.
This means that we have a $(1-1/e-\varepsilon,1-1/|E|)$-approximation
	oracle.

In the CMAB framework, we do not know the activation probabilities of edges
	and want to learn them during repeated seed selections while maximizing
	overall reward.
Each edge in $E$ is considered as a base arm, and
	a super arm in this setting is the set $E_S$ of edges incident to the seed set $S$.
Note that these edges will be deterministically triggered, but other edges not in $E_S$ may also be triggered, and the reward
	is related to all the triggered arms.
Therefore, this is an instance where arms may be probabilistically triggered, and thus $p_i<1$ for some $i\in E$.

It is straightforward to see that the expected reward function is still a function of probabilities on all edges, and
	the monotonicity holds.
However, bounded smoothness property is nontrivial to argue, as we will show in the following lemma.

\begin{mylem}
The social influence maximization instance satisfies the bounded smoothness property with bounded smoothness function $f(x)=|E||V|x$. \label{lem:social}
\end{mylem}
\begin{proof}
For the social influence maximization bandit, the expectation vector $\bp$ is the vector of all probabilities on all edges.
For a seed set $S\subseteq V$, the corresponding super arm is the set $E_S$ of edges incident to vertices in $S$.
Without loss of generality, we assume that for any edge $i\in E$, its probability $\mu_i>0$.
Then for super arm $E_S$, the set of base arms that can be triggered by $E_S$, denoted as $\trig{E}_S$, is exactly the set of edges reachable from seed set $S$
	(an edge $(u,v)$ reachable from a set $S$ means its starting vertex $u$ is reachable from $S$).
By Definition~\ref{def:assumptions}, to show bounded smoothness with bounded smoothness function
	$f(x)=|E||V|x$, we need to show that for any two expectation vectors $\bp$ and $\bp'$ and for any $\Lambda > 0$,
	we have $|r_{\bp}(E_S) - r_{\bp'}(E_S)| \leq f(\Lambda)$ if
	$\max_{i\in \trig{E}_S}|\mu_i - \mu_i'|\le \Lambda$. 

Since we know that monotonicity holds, it is sufficient to assume that for all $i\in \trig{E}_S$, $\mu_i = \mu'_i + \Lambda$.
This is because without loss of generality, we can assume $r_{\bp}(E_S) \ge r_{\bp'}(E_S)$, and if $\mu_i < \mu'_i + \Lambda$ we can increase $\mu_i$ and decrease $\mu'_i$ such that $\mu_i = \mu'_i + \Lambda$, 
	and this only increase the gap between $r_{\bp}(E_S)$ and $r_{\bp'}(E_S)$.
Thus, henceforth let us assume that $i\in \trig{E}_S$, $\mu_i = \mu'_i + \Lambda$.

Starting from $\bp'$, we take one edge $i_1$ in $\trig{E}_S$, and increase $\mu'_{i_1}$ to $\mu'_{i_1}+\Lambda = \mu_{i_1}$ to get a new vection $\bp^{(1)}$.
Suppose the edge $i_1$ is $(u_1, v_1)$.
Comparing $\bp'$ with $\bp^{(1)}$, the only difference is that the probability on edge $(u_1, v_1)$ increases by $\Lambda$.
For the influence spread of seed set $S$, the above change increases the activation probability of $v_1$ and every node reachable from $v_1$ by at most $\Lambda$. Thus the total increase of influence spread
	is at most $|V| \Lambda$.
Then we select the second edge $i_2$ in $\trig{E}_S$ and increases its probability by $\Lambda$.
By the same argument, the influence spread increases at most $|V|\Lambda$.
Repeating the above process, after selecting all edges in $\trig{E}_S$, we obtain probability vector $\bp^{(s)}$ where $s = |\trig{E}_S|$, and the increase in influence spread
	is at most $s|V|\Lambda$.
Comparing vector $\bp^{(s)}$ with $\bp$, they are the same on all edges in $\trig{E}_S$, and may only differ in the rest of edges. However, since the rest of edges cannot be reachable from $S$, their difference
	will not affect the influence spread of $S$.
Therefore, we know that the difference between influence spread $r_{\bp}(E_S)$ and $r_{\bp'}(E_S)$ is at most $s|V|\Lambda \le |E||V| \Lambda$.
This concludes that if we use function $f(x)=|E||V|x$, the bounded smoothness property holds.
\end{proof}

\paragraph{Remark.}
In Section 4.2 of~\citep{CWY13}, we made a claim that social influence maximization bandit satisfies the bounded smoothness property (with function $f(x)=|E||V|x$) that does not consider probabilistically triggered
	arms, that is, it satisfies the property that for any two expectation vectors $\bp$ and $\bp'$ and for any $\Lambda > 0$,
		$|r_{\bp}(E_S) - r_{\bp'}(E_S)| \leq f(\Lambda)$ if $\max_{i\in E_S}|\mu_i - \mu_i'|\le \Lambda$.
This claim is incorrect.
For example, all edges in $E_S$ could have the same probability (and thus we could have $\Lambda$ to be arbitrarily small), but other edges reachable from $E_S$ have different probabilities, and thus
	the gap between $r_{\bp}(E_S)$ and $r_{\bp'}(E_S)$ will not be arbitrarily small and cannot be bounded by $f(\Lambda)$ for any continuous $f$ tending to zero when $\Lambda$ tends to zero.

\chgred{With $f(x)=|E||V|x$, we have $\ell_n(\Delta,p)= \max\left (\frac{12\cdot \ln  n}{(f^{-1}(\Delta))^2\cdot p}, 
\frac{24\cdot \ln  n}{p} \right ) = \max \left (\frac{12 |V|^2|E|^2 \ln  n}{\Delta^2\cdot p}, 
\frac{24\cdot \ln  n}{p} \right )$.
Since $\Delta$ is at most $\Delta_{\max}$ in the regret bound and $\Delta_{\max}\le |V|$, it is clear that we have
	$\ell_n(\Delta,p)= \frac{12 |V|^2|E|^2 \ln  n}{\Delta^2\cdot p}$.
Then applying Theorem \ref{thm:cucb individual},  we know that the distribution-dependent 
	$(1-1/e-\varepsilon,1-1/|E|)$-approximation regret bound of the CUCB algorithm on influence maximization is:}
\[
\sum_{i\in E, K_i>0} \frac{24\cdot |V|^2|E|^2\cdot \ln n}{\Delta^i_{\min} \cdot p_i}
 \chgred{+\left(\frac{\pi^2}{2}+1\right)\cdot |E| \cdot \Delta_{\max}}.
\]
%
%
\chgred{With Theorem~\ref{thm:worstbound} (and further using $\ell_n(\Delta,p)= \frac{12 |V|^2|E|^2 \ln  n}{\Delta^2\cdot p}$
	instead of the relaxed $\ell_n(\Delta,p)= \frac{12 |V|^2|E|^2 \ln  n}{\Delta^2\cdot p} + \frac{24\cdot \ln  n}{p}$
	as in the proof of Theorem~\ref{thm:worstbound}), we obtain the distribution-independent bound:}
\[
|V|\sqrt{\frac{48|E|^3 n \ln n}{p^*} } +
 \chgred{\left(\frac{\pi^2}{2}+1\right)\cdot |E| \cdot \Delta_{\max}}.
\]

\vspace{\sectionspace}
\section{Conclusion}
\label{sec:conclude}
In this paper, we propose the first general stochastic CMAB framework that accommodates
	a large class of nonlinear reward functions
	among combinatorial and stochastic arms, and it even accommodates probabilistically
	triggered arms such as what occurs in the viral marketing application.
We provide CUCB algorithm with tight analysis on its distribution-dependent
	and distribution-independent regret bounds and applications to new practical
	combinatorial bandit problems.

There are many possible future directions from this work.
One may study the CMAB problems with Markovian outcome
	distributions on arms, or the restless version of CMAB, in which
	the states of arms continue to evolve even if they are not
	played.
\chgins{
Another direction is to investigate if some of the results in this paper are tight or
	can be further improved.
For example, for the nonlinear bounded smoothness function
	$f(x) = \gamma \cdot x^\omega$ with $\omega < 1$, if our bound in Theorem~\ref{thm:worstbound}
	is tight or can be improved; and
	for the case of probabilistic triggering, if the regret bound dependency on 
	$1/p_i$ is necessary.
For the latter case, one may also look into improvement specifically for the influence
	maximization application.
}
\chgins{For nonlinear reward functions, currently we assume that the expected reward is a function
	of the expectation vector of base arms.
One may also look into the more general cases where the expected reward depends not only on the expected outcomes
	of base arms.}


\bibliography{singlebib}
\bibliographystyle{icml2014}



\end{document}